\documentclass[11pt, a4paper]{article}
\usepackage[final]{acl}

\usepackage[utf8]{inputenc}
\usepackage{microtype}
\usepackage{inconsolata}
\usepackage{times}
\usepackage{latexsym}
\usepackage{enumitem}
\usepackage{multicol}
\usepackage{multirow}
\usepackage{amsmath, amsfonts}
\usepackage{amsthm}
\usepackage{bm}
\usepackage{tabularx}
\usepackage{graphicx}
\usepackage[linesnumbered,ruled,vlined]{algorithm2e}
\usepackage{caption}
\usepackage[list=true]{subcaption}
\usepackage{bbm}
\usepackage{float}
\usepackage{colortbl}
\usepackage{balance}
\usepackage{booktabs}
\usepackage{tcolorbox}
\usepackage{hyperref}
\usepackage{url}

\usepackage{enumitem}
\usepackage{arydshln} 
\setlistdepth{9}

\setlist[itemize,1]{label=$\bullet$}
\setlist[itemize,2]{label=$\bullet$}
\setlist[itemize,3]{label=$\circ$}
\setlist[itemize,4]{label=$\ast$}
\setlist[itemize,5]{label=$\diamond$}
\setlist[itemize,6]{label=$\bullet$}
\setlist[itemize,7]{label=$\bullet$}
\setlist[itemize,8]{label=$\bullet$}
\setlist[itemize,9]{label=$\bullet$}

\renewlist{itemize}{itemize}{9}

\newcommand{\vpara}[1]{\vspace{0.07in}\noindent\textbf{#1 }}
\renewcommand{\vec}[1]{\mathbf{#1}}
\newcommand{\hide}[1]{}
\def\hide#1{\textcolor{white}{#1}}

\title{Detecting and Explaining Fake News Short Videos with Multimodal Content and Real-World Evidence}

\author{
  \textbf{Yifeng Luo}\textsuperscript{1,2}
  \quad
  \textbf{Yupeng Li}\textsuperscript{1,2}
  \thanks{Yifeng Luo conducted this work under the supervision of Yupeng Li and Liang Lan. Yupeng Li (\url{ivanypli@gmail.com}) is the corresponding author.}
  \quad
  \textbf{Ming Tang}\textsuperscript{4}
  \quad
  \textbf{Jianxiong Guo}\textsuperscript{5}
  \quad
  \textbf{Liang Lan}\textsuperscript{1,3}
  \\
  \normalfont
  \textsuperscript{1}Department of Interactive Media;
  \textsuperscript{2}AI and Social Good Lab, AI Media Centre;
  \textsuperscript{3}AI Media Centre\\
  Hong Kong Baptist University\\
  \textsuperscript{4}Department of Computer Science and Engineering,
  Southern University of Science and Technology\\
  \textsuperscript{5}Institute of AI and Future Networks,
  Beijing Normal University\\
}

\begin{document}
\maketitle

\begin{abstract}
Short-video platforms have become a primary news source for the public, which has also enabled the widespread dissemination of fake news videos. We study the task of fake news video detection and explanation (FNVDE). Existing methods face two critical limitations. 
First, commonly used frame selection strategies may omit veracity-relevant cues or provide insufficient temporal context for understanding news videos.
Second, prior methods neglect either multimodal understanding or evidence retrieval. To address these limitations, we propose NVKE-CEI, a unified system that integrates a news video keyframes extraction method (NVKE) and an FNVDE framework leveraging both content and evidence information (CEI). 
NVKE selects keyframes based on chronological changes in combined visual and OCR-text similarity.
CEI employs two specialized LLM-based fact checkers (content-based and evidence-based) whose outputs are fused by a lightweight judge model. Extensive experiments show that NVKE-CEI outperforms state-of-the-art baselines while generating high-quality content-grounded explanations.
\end{abstract}

\section{Introduction}
\label{Sec:Intr}

\begin{figure*}[!t]
\centering
\includegraphics[scale=2.12]{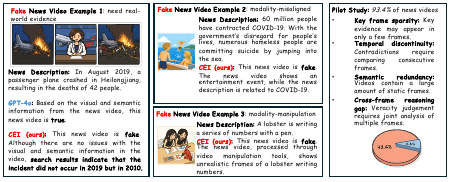}
\caption{In fake news video example 1, the opening and concluding segments both depict the accident scene with minimal visual variation but convey different information: the opening introduces the accident, whereas the concluding segment shows rescue operations. Density-based keyframe extraction methods may miss these semantically distinct frames due to their visual similarity.
Furthermore, verifying the veracity requires real-world evidence. Fake news video example 2 (resp.~3) illustrates the category of modality-misaligned (resp.~modality-manipulated) fake news video. In the pilot study, we identify four frame-selection challenges in the FakeSV dataset.}
\label{F1}
\end{figure*}

Short-video platforms have become a major channel for daily news consumption and for the dissemination of fake news videos~\cite{wang2021seeing}, i.e., videos containing false or misleading information~\cite{videosurvey, baseline1, baseline2, luo2026cheap}.
Recent studies have found that
video-based misinformation tends to spread faster and is perceived as more trustworthy by users than the same misinformation when presented in text or image formats, posing significant risks to public opinion and social good~\cite{danger1, sundar2021seeing, li2024mcfend}.
To combat this,
recent studies~\cite{hong2025following, yan2025debunk, zhang2025fact}
have utilized
the pretrained knowledge and generative capabilities of Large Language Models (LLMs\footnote{Throughout this paper, we use ``LLMs'' as an umbrella term for large foundation models, including text-only, vision-language, and video-language models.})~\cite{gong2023multimodal, achiam2023gpt} to detect fake news videos. 

In this paper, we consider the task of fake news video detection and explanation (FNVDE).
Given a news video, FNVDE aims to predict its veracity label and generate a news content-grounded explanation, i.e., an explanation of why the news video is real or fake based on its news content~\cite{cheng2021socially, wang2023explainable, wang2024explainable, khaliq2024ragar, he2026debating}.
This differs from post-hoc interpretation methods that highlight influential regions~\cite{wu2024interpretable} and from supervised explanation-generation methods that require human-annotated explanations~\cite{chen2025multimodal}.
Existing LLM-based methods either summarize news videos into text and leverage pretrained LLM knowledge~\cite{hong2025following, yan2025debunk}, or fine-tune LLMs with annotated text-video pairs~\cite{zhang2025fact}. However, the hallucination problem~\cite{ji2023towards, perkovic2024hallucinations} persists, as the
LLM reasoning process
is not grounded in real-world evidence, leading to incorrect veracity predictions~\cite{wang2024mfc, hu2024bad}. 
To mitigate hallucinations, recent methods ground LLM reasoning in real-world evidence by either training models for Deep Research~\cite{team2025tongyi} or prompting them to use retrieval-augmented generation (RAG)~\cite{asai2023self, lewis2020retrieval, he2026novel}.
These methods generate search queries by extracting keywords from news~\cite{liu2025detect} or providing query examples to LLMs~\cite{singhal2024evidence}, then submit them to web search engines and integrate retrieved evidence to predict news veracity. 

Despite this progress, applying LLM-based methods to FNVDE presents several challenges.
First, frame selection determines what visual information is available to an LLM for understanding a news video.
Directly inputting all frames of a news video into the LLM context window is costly and often redundant.
Existing LLM-based FNVDE methods rely on random frame sampling~\cite{hong2025following, yan2025debunk, zhang2025fact}, 
which may omit briefly occurring but veracity-relevant cues or fail to preserve sufficient temporal context, thereby limiting the LLM's ability to form a clear understanding of the news video.
Existing LLM-based video understanding methods~\cite{zhou2024streaming, jin2024chat} 
employ clustering-based visual compression, but these general strategies are not specifically designed to select news video frames according to joint changes in visual and OCR-text content over time. They may therefore underrepresent visually similar frames that convey different information at different time points.
Our pilot study (see Appendix~\ref{Pilot}) on the FakeSV dataset~\cite{baseline1} finds that $93.4\%$ of the analyzed news videos 
exhibit at least one of four frame-selection challenges, highlighting the difficulty of preserving both semantically important cues and sufficient temporal context in a limited frame set (see Fig.~\ref{F1}).
Second, both multimodal understanding and evidence retrieval are important. Effective FNVDE methods require both multimodal understanding to identify manipulation (e.g., modality misalignment or content fabrication) and real-world evidence retrieval to ground reasoning and avoid hallucinations (see Fig.~\ref{F1}). However, existing LLM-based methods 
either focus solely on multimodal content analysis~\cite{hong2025following, yan2025debunk, zhang2025fact} or exclusively on external evidence retrieval~\cite{wang2024explainable, choi2024fact, khaliq2024ragar, liu2025detect}. This singular focus leads to limitations when handling diverse fake news types.
Moreover, evidence-based methods employing RAG assume web-retrieved evidence is reliable, yet such evidence can be noisy and contain misleading information~\cite{ge2025resolving, luo2024message, he2026fact2fiction}, further complicating the FNVDE task.

To address these limitations, we propose \textsc{NVKE-CEI}, which contains \textsc{NVKE}, a news video keyframes extraction method, and \textsc{CEI}, a novel FNVDE framework that leverages both content and evidence information for veracity prediction and content-grounded explanation generation.
CEI employs two specialized LLM-based fact checkers, namely a content-based LLM and an evidence-based LLM, whose complete reasoning outputs constitute the content-grounded explanation; a lightweight judge model integrates their outputs to produce the final prediction. We propose NVKE to address the first limitation by selecting keyframes according to the magnitude of chronological changes in visual-text similarity, with the goal of preserving semantic changes and temporal context in news videos.
To address the second limitation, CEI utilizes LLM role-playing~\cite{wang2024rolellm, shanahan2023role} to handle diverse fake news video types via a decoupled architecture.
We argue that a decoupled architecture is essential for FNVDE rather than using a single LLM. A single LLM may be affected by knowledge cutoffs~\cite{cheng2024dated} and confirmation bias~\cite{wan2025unveiling}. 
When the same LLM analyzes content and retrieves evidence jointly, its initial assessment may influence how it interprets subsequently retrieved evidence~\cite{wang2025truth}.
CEI therefore adopts a dual-perspective fact-checking process. 
The content-based LLM focuses on internal inconsistencies and provides content-driven predictions, while the evidence-based LLM focuses primarily on external verification. 
The evidence-based LLM, inspired by human fact-checking processes~\cite{vlachos2014fact, brandtzaeg2018journalists}, employs reasoning and acting steps~\cite{react} to generate search queries, gather real-world evidence, and make evidence-grounded predictions. 
This decoupling keeps the two reasoning streams distinct before fusion, allowing the lightweight judge model to combine their signals through text feature attention and veracity score fusion to produce the final prediction.
We address retrieval noise at two stages: the query regeneration module refines search queries before retrieval, while the text feature attention module allows the lightweight judge to adaptively weight retrieved evidence alongside the other textual features. The main contributions are summarized as follows:
\begin{itemize}[leftmargin=*, labelindent=0pt]
    \item We propose NVKE to select keyframes for the LLM context window by tracking chronological changes in visual-text similarity, aiming to preserve content changes and temporal context while reducing redundancy.
    \item We combine multimodal content analysis and evidence retrieval through specialized LLM checkers, employing query regeneration in evidence-based LLM and attention mechanisms in a lightweight judge model to handle noisy web evidence effectively.
    \item Extensive experiments on FakeSV and FakeTT with thirteen baselines demonstrate that NVKE-CEI achieves state-of-the-art (SOTA) performance. Compared with the second-best results, NVKE-CEI achieves an increase of $0.0071$ (resp.~$0.0234$) in Acc. and $0.0002$ (resp.~$0.0190$) in Macro $\vec{F}_1$ score on FakeSV (resp.~FakeTT)\footnote{Source code is available at \url{https://github.com/TrustworthyComp/CEI}.}.
\end{itemize}

\begin{figure*}[!t]
\centering 
\includegraphics[scale=1.85]{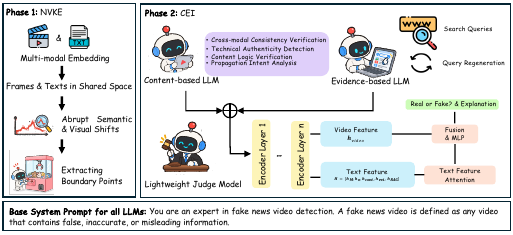}
\caption{Overview of our proposed NVKE-CEI.}
\label{overview}
\end{figure*}

\section{NVKE-CEI}
In this section, we begin by formally defining the FNVDE task. We propose NVKE to extract the news video keyframes by considering both semantic and temporal information. Building upon the NVKE, we introduce the CEI framework. The CEI framework consists of two specialized LLM fact checkers and one lightweight judge model.
The overview of CEI is presented in Fig.~\ref{overview}.

\subsection{Problem Setting}
We denote a news video as $\mathcal{V} = \{\vec{F}, M\}$ with frames $\vec{F} = \{\vec{f}_1, \ldots, \vec{f}_n\}$ and text material $M$. Here, the text material $M$ is the concatenation of the title, description, and audio transcript of the news video. 
Each news video $\mathcal{V}$ has a binary veracity label $y \in \mathcal{Y}$, where $\mathcal{Y}=\{\mathrm{real},\mathrm{fake}\}$.
The task of FNVDE is to predict the veracity label $\hat{y}$ and the news content-grounded explanation $e$.

\subsection{NVKE: News Video Keyframes Extraction}

\vpara{Visual Text Embedding and Similarities.} Given a news video $\mathcal{V}$ with frames $\vec{F} = \{\vec{f}_1, \ldots, \vec{f}_n\}$, we extract optical characters (OC) $\vec{O} = \{\vec{o_1}, \ldots, \vec{o_n}\}$ using the OCR tool, $\vec{O} = \text{Tool}_\text{ocr}(\vec{F})$. 
To capture both visual and semantic information, we use CLIP \cite{radford2021learning} encoders to embed them into a shared space: $\vec{H}_\vec{F} = \text{CLIP}_\text{visual}(\vec{F})$ and $\vec{H}_\vec{O} = \text{CLIP}_\text{text}(\vec{O})$, where $\vec{H}_\vec{F} = \{\vec{h}_\vec{f}^1, \ldots, \vec{h}_\vec{f}^n\}$ and $\vec{H}_\vec{O} = \{\vec{h}_\vec{o}^1, \ldots, \vec{h}_\vec{o}^n\}$ are the visual and semantic embeddings respectively. 
We compute the cosine similarity between consecutive frames and OCs, i.e.,
visual similarities $\vec{S}_\vec{F} = \{s_\vec{F}^1, \ldots, s_\vec{F}^{n-1}\}$ and semantic similarities $\vec{S}_\vec{O} = \{s_\vec{O}^1, \ldots, s_\vec{O}^{n-1}\}$ as $s_\vec{F}^i = \frac{\vec{h}_\vec{f}^i \cdot \vec{h}_\vec{f}^{i+1}}{||\vec{h}_\vec{f}^i|| ||\vec{h}_\vec{f}^{i+1}||}$ and $s_\vec{O}^i = \frac{\vec{h}_\vec{o}^i \cdot \vec{h}_\vec{o}^{i+1}}{||\vec{h}_\vec{o}^i|| ||\vec{h}_\vec{o}^{i+1}||}$ for $i \in \{1, 2, \ldots, n-1\}$. The overall visual text similarities $\vec{S} = \{s_1, \ldots, s_{n-1}\}$ are: $\vec{S} = \alpha \vec{S}_\vec{F} + (1-\alpha) \vec{S}_\vec{O}$ with hyper-parameter $\alpha \in [0, 1]$.

\vpara{Keyframes and OC Extraction.} To capture temporal intensity of content variation, we compute the magnitude of similarity changes $\Delta S = \{\Delta s_1, \ldots, \Delta s_{n-2}\}$, where $\Delta s_i = |s_{i+1} - s_i|$ for $i \in \{1, 2, \ldots, n-2\}$. 
 We sort all change points in descending order of $\Delta s_i$, where each $\Delta s_i$ corresponds to three consecutive frames $\vec{f}_i$, $\vec{f}_{i+1}$, and $\vec{f}_{i+2}$ and their aligned OCR texts.
Starting from the highest-ranked change point, we add its two boundary frames $\vec{f}_i$ and $\vec{f}_{i+2}$ and continue down the ranking until at least $2K$ distinct frames have been collected or no change points remain.
The selected keyframes and OCs are denoted as $\vec{F}_\text{key} = \{\vec{f}_1, \ldots, \vec{f}_{2K}\}$ and $\vec{O}_\text{key} =\{\vec{o}_1, \ldots, \vec{o}_{2K}\}$.

\subsection{CEI: FNVDE Framework with Content and Evidence Information}

\vpara{Base System Prompt and Structured Output.} The two LLM checkers share a common base prompt that defines fake news videos and explains the task of FNVDE before assigning their specialized roles. The base system prompt is shown in Fig.~\ref{overview}.
To extract well-defined components from the LLM fact-checkers' responses, we employ structured output\footnote{\url{https://openai.com/index/introducing-structured-outputs-in-the-api/}}.
For the content-based LLM, which operates in a single turn, the output includes a reasoning process $\mathcal{R}_\text{cont.}$ that justifies the predicted confidence score $\hat{\vec{y}}_\text{cont.}$.
For the evidence-based LLM, which follows an iterative ReAct framework, the output structure depends on the dialogue stage. In intermediate turns, the LLM outputs a reasoning process $\mathcal{R}_\text{evi.}$ explaining its information needs and generates search queries $\vec{q}$ accordingly. In the final turn, after synthesizing retrieved evidence, it outputs $\mathcal{R}_\text{evi.}$ justifying the confidence score $\hat{\vec{y}}_\text{evi.}$. Both confidence scores represent the estimated probability of the fake class.

\subsubsection{Content-based LLM}

\vpara{Content-based LLM Fake News Video Checker.} The content-based LLM checker examines semantic and syntactic manipulations within news videos through comprehensive multimodal analysis. To enhance its analytical capability, we employ role-playing~\cite{wang2024rolellm, shanahan2023role} by instructing the LLM to act as \textit{a seasoned fake news video detection expert with extensive expertise in media literacy, multimodal content analysis, and fact-checking methodologies}. Under this role, the LLM performs systematic verification across four core dimensions: (1) Cross-modal consistency verification analyzes potential contradictions between textual descriptions, visual content, and audio information; (2) Technical authenticity detection examines video frames for evidence of artificial modification, splicing, or fabrication; (3) Content logic verification assesses whether video content contains factual errors or logical fallacies; (4) Propagation intent analysis identifies inflammatory or manipulative expressions that may mislead the audience. 
Given a news video $\mathcal{V}$, we utilize structured output to simultaneously obtain the reasoning process $\mathcal{R}_\text{cont.}$ and confidence score $\hat{\vec{y}}_\text{cont.}$ from the content-based LLM $\text{LLM}_\text{cont.}(\cdot)$,
$\hat{\vec{y}}_\text{cont.}, \mathcal{R}_\text{cont.} = \text{LLM}_\text{cont.}(\vec{F}_\text{key}, \vec{O}_\text{key}, M)$.
The detailed prompt for the content-based LLM is presented in Appendix~\ref{P:Content}.

\subsubsection{Evidence-based LLM}

While the content-based LLM addresses internal manipulations within news videos, many fake news videos contain factually incorrect claims that require real-world evidence to verify. To address this, we design an evidence-based LLM that grounds its reasoning in external evidence retrieved from the web. The checker operates through an iterative ReAct framework consisting of three integrated components: (1) the evidence-based LLM employs reasoning and acting steps to analyze video content, generate search queries, and use the retrieved evidence for veracity prediction~\cite{react}; (2) the query regeneration module refines initial LLM-generated queries to improve evidence quality, addressing the problem that the retrieved evidence may contain noisy information at the retrieval stage; and (3) the RAG pipeline retrieves real-world evidence from web search engines. 

\vpara{Evidence-based LLM with ReAct Framework.} To handle fake news videos containing factually incorrect claims, the evidence-based LLM checker verifies information through real-world evidence retrieval. Inspired by human fact-checking processes~\cite{guo2022survey, brandtzaeg2018journalists}, we employ the ReAct framework~\cite{react}, where the LLM alternates between reasoning (understanding video content, formulating information needs, and assessing veracity) and acting (generating search queries to retrieve evidence). This iterative process continues until the LLM gathers sufficient evidence to produce a confidence score for its veracity prediction. Specifically, the evidence-based LLM operates through two types of reasoning steps. First, in the query generation reasoning step, the LLM analyzes the news video and outputs both its reasoning process $\mathcal{R}_\text{evi.}$ and search queries $\vec{q}$. Initially, this occurs by understanding the video content,
$\vec{q}, \mathcal{R}_\text{evi.} = \text{LLM}_\text{evi.}(\vec{F}_\text{key}, \vec{O}_\text{key}, M)$. 
In subsequent iterations, if additional information is needed, the LLM generates new queries based on previously retrieved evidence $\mathcal{R}_\text{RAG}$,
$\vec{q}, \mathcal{R}_\text{evi.} = \text{LLM}_\text{evi.}(\mathcal{R}_\text{RAG})$.
Second, in the veracity prediction reasoning step, once the LLM determines that sufficient evidence has been gathered, it uses the retrieved information to output its confidence score $\hat{\vec{y}}_\text{evi.}$ and final reasoning process $\mathcal{R}_\text{evi.}$:
$\hat{\vec{y}}_\text{evi.}, \mathcal{R}_\text{evi.} = \text{LLM}_\text{evi.}(\mathcal{R}_\text{RAG})$. 
We denote $\vec{q}^\tau = \{\vec{q}^1, {\vec{q}^\prime}^1, \ldots, \vec{q}^{T-1}, {\vec{q}^\prime}^{T-1}\}$ as the complete sequence of queries, $\mathcal{R}^\tau_\text{evi.} = \{\mathcal{R}^1_\text{evi.}, \ldots, \mathcal{R}^T_\text{evi.}\}$ as the complete sequence of reasoning process, and $\mathcal{R}^\tau_\text{RAG} = \{\mathcal{R}_\text{RAG}^1, \ldots, \mathcal{R}_\text{RAG}^{T-1}\}$ as the complete sequence of retrieved evidence. 
The complete output trajectory is denoted as $\tau = \{\vec{q}^1, \mathcal{R}_\text{evi.}^1, {\vec{q}^\prime}^1, \mathcal{R}_\text{RAG}^1, \ldots, \mathcal{R}_\text{RAG}^{T-1}, \mathcal{R}_\text{evi.}^T, \hat{\vec{y}}_\text{evi.}\}$, where $t \in \{1, \ldots, T\}$ indexes an iterative step and $T$ is the total number of iterative steps; query generation and evidence retrieval occur for $t \in \{1, \ldots, T-1\}$, while the final veracity prediction is produced at $t=T$.
The detailed prompt for the evidence-based LLM is presented in Appendix~\ref{P:Evidence}.

\vpara{Query Regeneration.} Initial LLM-generated queries may suffer from two critical issues that lead to low-quality evidence retrieval: they may contain redundant contextual information from the video description, and they may lack the specificity needed to retrieve precise, relevant evidence. To address this at the retrieval stage, we introduce a query regeneration module that refines each generated query $\vec{q}$ into a more focused version $\vec{q}^\prime$:
$\vec{q}^\prime = \text{LLM}_\text{query}(\vec{q})$, 
where $\text{LLM}_\text{query}(\cdot)$ evaluates queries based on specificity and direct relevance to core factual claims in the video. By generating more targeted queries, this module improves the quality of retrieved evidence. The detailed prompt is presented in Appendix~\ref{P:QueryRegen}.

\vpara{RAG Pipeline.} To retrieve real-world evidence, we implement a RAG pipeline using a web search engine as the retriever~\cite{welz2024enhancing, borgeaud2022improving}. Given a refined search query $\vec{q}^\prime$, the RAG retriever $\text{RAG}(\cdot)$ returns the $K_\text{RAG}$ most relevant webpage snippets:
$\mathcal{R}_\text{RAG} = \text{RAG}(\vec{q}^\prime)$.
These retrieved snippets provide real-world evidence that the evidence-based LLM uses to make grounded veracity predictions, avoiding hallucinations that may occur when relying solely on the LLM's pretrained knowledge.

\subsection{Lightweight Judge Model}
We employ a lightweight neural network as the final judge module instead of the meta-judge LLM for two primary reasons. First, while LLMs possess strong general reasoning capabilities, FNVDE often requires capturing subtle, dataset-specific distribution biases (e.g., specific linguistic patterns or platform-specific metadata correlations). A lightweight model trained via supervised learning can effectively capture these specific patterns, often outperforming a general-purpose LLM operating in a zero-shot or few-shot setting. 
Second, the analyses generated by the content-based and evidence-based agents are typically lengthy and token-heavy. Concatenating these long contexts and feeding them back into a commercial LLM API (like GPT-4o) for a final verdict would incur prohibitive token costs, especially for large-scale datasets. In contrast, the lightweight judge processes the fused representations with negligible computational overhead and zero additional API costs, making the framework more practical for real-world deployment.
The judge model consists of four key components: (1) \textit{feature encoding} processes multiple text sources and video frames using a small model; (2) \textit{text feature attention} employs self-attention to fuse heterogeneous text features while filtering noisy evidence; (3) \textit{multimodal fusion} integrates text and video features through concatenation; (4) \textit{veracity score fusion} incorporates LLM confidence scores as prior knowledge.

\vpara{Feature Encoding.} The judge model processes five text features (text material $M$, key OCs $\vec{O}_\text{key}$, content-based reasoning $\mathcal{R}_\text{cont.}$, evidence-based reasoning process $\mathcal{R}^\tau_\text{evi.}$, retrieved evidence $\mathcal{R}^\tau_\text{RAG}$) and one visual feature (keyframes $\vec{F}_\text{key}$). We employ CLIP encoders~\cite{radford2021learning} and MLPs to project them into a shared hidden space via
$
\vec{h}_i = \text{MLP}_\text{text}(\text{CLIP}_\text{text}(\cdot)) + \vec{b}_i, 
$
and 
$
\vec{h}_\text{video} = \text{AvgPool}(\text{MLP}_\text{visual}(\text{CLIP}_\text{visual}(\vec{F}_\text{key})))
$
respectively, where $i \in \{M, \vec{O}, \text{cont.}, \text{evi.}, \text{RAG}\}$ and $\vec{b}_i$ are learnable biases.

\vpara{Text Feature Attention and Multimodal Fusion.} To integrate heterogeneous text features while filtering noisy evidence, we apply self-attention over stacked text features $\vec{H} = [\vec{h}_M; \vec{h}_{\vec{O}}; \vec{h}_\text{cont.}; \vec{h}_\text{evi.}; \vec{h}_\text{RAG}] \in \mathbb{R}^{5 \times d}$. The attended features are computed as:
$
\vec{H}_\text{attended} = \text{softmax}\left(\frac{\vec{H}\vec{Q}(\vec{H}\vec{K})^\top}{\sqrt{d}}\right) \vec{H}\vec{V},
$
where $\vec{Q}, \vec{K}, \vec{V} \in \mathbb{R}^{d \times d}$ are learnable projection matrices and $d$ is the hidden dimension. The fused text feature is obtained by $\vec{h}_\text{text} = \frac{1}{5}\sum_{i=1}^5 \vec{H}_\text{attended}[i]$, which adaptively down-weights noisy retrieved evidence. We then concatenate text and video features as $\vec{h}_\text{fused} = [\vec{h}_\text{text}; \vec{h}_\text{video}] \in \mathbb{R}^{2d}$, preserving information from both modalities while allowing the classifier to learn their interactions. 
The judge logits are obtained through a two-layer MLP classifier:
$\hat{\vec{y}}_\text{judge} = \text{MLP}(\vec{h}_\text{fused}) \in \mathbb{R}^2$, where the two entries correspond to the \texttt{real} and \texttt{fake} classes.

\vpara{Veracity Score Fusion.} To leverage the complementary strengths of both LLM checkers, we introduce a logit-level veracity score fusion mechanism that injects LLM confidence scores as prior knowledge into the model's predictions. Given the confidence scores $\hat{\vec{y}}_\text{cont.}$ and $\hat{\vec{y}}_\text{evi.}$ from the content-based and evidence-based LLMs respectively, 
we transform their fake-class probabilities into log-odds so that they can be combined additively with the judge logits.
Specifically, we compute the logit of the fake news probability from each LLM:
$
\ell_\text{cont.} = \log\left(\frac{p_\text{cont.}}{1-p_\text{cont.}}\right), \ell_\text{evi.} = \log\left(\frac{p_\text{evi.}}{1-p_\text{evi.}}\right),
$
where $p_\text{cont.} = \hat{\vec{y}}_\text{cont.}[\texttt{fake}]$ and $p_\text{evi.} = \hat{\vec{y}}_\text{evi.}[\texttt{fake}]$ are clamped to $[\epsilon, 1-\epsilon]$ with $\epsilon=10^{-6}$.
The combined prior is $\ell_\text{prior} = 0.5 \cdot \ell_\text{cont.} + 0.5 \cdot \ell_\text{evi.}$, which is injected into the model's fake class logit, $
\hat{\vec{y}}_\text{final} = \hat{\vec{y}}_\text{judge} + [0, \gamma \cdot \ell_\text{prior}]$.
Here, $\gamma$ is a learnable scale parameter initialized to 1.0, enabling the model to balance its predictions with LLM priors automatically.
The judge model is trained using standard cross-entropy loss with label smoothing ($\epsilon_\text{smooth}=0.1$). 
The class corresponding to the larger entry in $\hat{\vec{y}}_\text{final}$ is returned as the final veracity label $\hat{y}$.
The content-grounded explanation $e$ consists of the content-based reasoning $\mathcal{R}_\text{cont.}$ and the complete sequence of evidence-based reasoning processes $\mathcal{R}^\tau_\text{evi.}$.

\begin{table*}[!t]
\centering
\caption{Performance compared with the baselines.}
\resizebox{1\textwidth}{!}{
\begin{tabular}{cccccccccc}
\toprule
 & \multicolumn{4}{c}{\textbf{FakeSV} \cite{baseline1}} & & \multicolumn{4}{c}{\textbf{FakeTT} \cite{baseline2}} \\
\cmidrule{2-5} \cmidrule{7-10}
Method & Acc. & Macro $\vec{F}_1$ & Prec. & Rec. & & Acc. & Macro $\vec{F}_1$ & Prec. & Rec. \\
\midrule
\textbf{GPT-4o}~\cite{achiam2023gpt} & 0.6518 & 0.6373 & 0.7527 & 0.6865 & & 0.7090 & 0.6509 & 0.6633 & 0.6448 \\ 
\textbf{DeepSeek-V3}~\cite{deepseek} & 0.6974 & 0.6920 & 0.7500 & 0.7043 & & 0.7224 & 0.6789 & 0.6845 & 0.5354\\
\textbf{Qwen3-vl-plus}~\cite{qwen} & 0.5657 & 0.5516 & 0.6300 & 0.6007 & & 0.6738 & 0.5131 & 0.6044 & 0.5409 \\
\textbf{Claude-3-5-sonnet} & 0.5918 & 0.5833 & 0.6387 & 0.6177 & & 0.6333 & 0.5990 & 0.5978 & 0.6011\\
\textbf{CoRAG}~\cite{khaliq2024ragar} & 0.3326 & 0.3292 & 0.6764 & 0.2189 & & 0.6720 & 0.6889 & 0.8204 & 0.5981 \\
\textbf{ReAct}~\cite{react} & 0.6084 & 0.6016 & 0.7625 & 0.5417 & & 0.6519 & 0.6643 & 0.7682 & 0.5904\\
\textbf{3MFact}~\cite{niu2025pioneering} & 	0.8122 & 0.8359 & 0.7970 & \textbf{0.8788} & &  0.7661 & 0.7920 & 0.7339 & \textbf{0.8601} \\
\textbf{FANVM}~\cite{choi2021using} & 0.7952 & 0.7881 & 0.7982 & 0.7788 & & 0.7180 &  0.7030 & 0.7023 & 0.7289 \\
\textbf{TikTec}~\cite{shang2021multimodal} & 0.7343 & 0.7326 & 0.7324 & 0.7349 & & 0.6622 & 0.6508 & 0.6584 & 0.6785\\
\textbf{FactR1}~\cite{zhang2025fact} & 0.7662 & 0.7478 & 0.7885 & 0.7111 & & 0.7444 &	0.7270 & 0.7925 & 0.6715 \\
\textbf{SVFEND}~\cite{baseline1} & 0.8086 & 0.8055 & 0.8018 & 0.8060 & & 0.7714 & 0.7563 & 0.7512 & 0.7756 \\
\textbf{FakingRecipe}~\cite{baseline2} & 0.8460 &  0.8434 & \underline{0.8585} & 0.8435 & & 0.7903 & 0.7770 & 0.7721 & 0.7980 \\
\textbf{ExMRD}~\cite{hong2025following} & \underline{0.8490} & \underline{0.8512} & 0.8286 & \underline{0.8771} & & \underline{0.8428} & \underline{0.8315} & \underline{0.8227} & 0.8519 \\
\midrule
\textbf{NVKE-CEI} (ours) & \textbf{0.8561} & \textbf{0.8514} & \textbf{0.8615} & 0.8471 & & \textbf{0.8662} & \textbf{0.8505} & \textbf{0.8473} & \underline{0.8541}\\
\bottomrule
\end{tabular}
\label{TAB:SUP}
}
\end{table*}

\section{Experiment}
\label{Sec.EXPALL}
In this section, we introduce the experiment settings, then evaluate NVKE-CEI in terms of detection performance and explainability. Additional experimental details and results are provided in Appendices~\ref{EXP:MOREEXP},~\ref{APP:LLMBack},~\ref{APP:Frame},~\ref{APP:MOREEXP}, and~\ref{Error}.

\subsection{Experiment Settings}
\vpara{Datasets and Evaluation Metrics.} 
In our experiments, we consider two widely used datasets in the literature on the detection of fake news videos, FakeSV~\cite{baseline1} and FakeTT~\cite{baseline2}. To evaluate the performance of our proposed fake news video detection method, we employ two evaluation metrics: accuracy (Acc.) and Macro $\vec{F}_1$ score. More details of the datasets and evaluation metrics can be found in Appendix~\ref{APP:dataset}.

\vpara{Baselines and Implementation Details.}
We evaluated the performance of our proposed fake news video detection framework with thirteen baselines, seven zero-shot LLM-based baselines, and six supervised training-based baselines. The seven zero-shot LLM-based baselines are \textbf{GPT-4o}~\cite{achiam2023gpt}, \textbf{DeepSeek-V3}~\cite{deepseek}, \textbf{Qwen3-vl-plus}~\cite{qwen}, \textbf{Claude-3-5-sonnet}\footnote{\url{https://www.anthropic.com/news/claude-3-5-sonnet/}}, \textbf{CoRAG}~\cite{khaliq2024ragar}, \textbf{ReAct}~\cite{react}, and \textbf{3MFact}~\cite{niu2025pioneering}. The six supervised training-based baselines are \textbf{FANVM}~\cite{choi2021using}, \textbf{TikTec}~\cite{shang2021multimodal}, \textbf{SVFEND}~\cite{baseline1}, \textbf{FakingRecipe}~\cite{baseline2}, \textbf{FactR1}~\cite{zhang2025fact}, and \textbf{ExMRD}~\cite{hong2025following}. The implementation details of the baselines can be found in Appendix~\ref{APP:baseline}, and the implementation details of our proposed NVKE-CEI framework can be found in Appendix~\ref{APP:CEIIMP}. 

\begin{table}[!t]
\centering
\caption{Ablation study of the proposed modules.}
\resizebox{0.48\textwidth}{!}{
\begin{tabular}{ccccc}
\toprule
Dataset & \multicolumn{2}{c}{FakeSV~\cite{baseline1}} & \multicolumn{2}{c}{FakeTT~\cite{baseline2}} \\
Variant & Acc. & Macro $\vec{F}_1$ & Acc. & Macro $\vec{F}_1$ \\
\midrule
w/o NVKE & 0.8210 & 0.8198 & 0.8395 & 0.8138 \\
w/o Content-based LLM & 0.8266 & 0.8254 & 0.8227 & 0.7994 \\
w/o Query Regeneration & 0.8395 & 0.8327 & 0.8428 & 0.8300 \\
w/o Evidence-based LLM & 0.7989 & 0.7985 & 0.8027 & 0.7493 \\
w/o Text Feature Attention & 0.8081 & 0.8081 & 0.7960 & 0.7880 \\
w/o Veracity Score Fusion & 0.8229 & 0.8227 & 0.8194 & 0.8051 \\
w/o Fine-tuning & 0.7968 & 0.7985 & 0.7543 & 0.7253 \\
\midrule
\textbf{NVKE-CEI} (ours) & \textbf{0.8561} & \textbf{0.8514} & \textbf{0.8662} & \textbf{0.8505} \\
\bottomrule
\end{tabular}
}
\label{TAB:AB}
\end{table}

\subsection{Experiments Results}
The detection performance of the baselines and our proposed NVKE-CEI framework is presented in Table~\ref{TAB:SUP}. To comprehensively evaluate the individual module contribution of the NVKE-CEI, we conduct extensive ablation studies, the results of which are detailed in Table~\ref{TAB:AB}. We use bolded (resp.~underlined) numbers to represent the best (resp.~second best) experiment results. Moreover, our proposed NVKE-CEI achieves the optimal trade-off between time efficiency and performance compared to other baselines (see Appendix~\ref{App:report}).

\vpara{Performance compared with baselines.} The NVKE-CEI framework achieves the best Acc. and Macro $\vec{F}_1$ scores on both FakeSV~\cite{baseline1} and FakeTT~\cite{baseline2} datasets. Compared with the second-best results, our proposed NVKE-CEI framework achieves an increase of $0.0071$ (resp.~$0.0234$) in Acc. and $0.0002$ (resp.~$0.0190$) Macro $\vec{F}_1$ score on FakeSV (resp.~FakeTT) dataset.
The superior performance of NVKE-CEI over all other baselines, particularly ExMRD~\cite{hong2025following} and FakingRecipe~\cite{baseline2}, demonstrates that our proposed NVKE-CEI framework has achieved SOTA performance.
We also present NVKE comparison experiments, case studies, and error analysis in the Appendices~\ref{APP:Frame} and~\ref{Error}.

\vpara{Ablation Studies.} 
Results in Table~\ref{TAB:AB} evaluate: (1) NVKE effectiveness, (2) LLM components (content-based LLM, query regeneration, evidence-based LLM), and (3) judge model components (text feature attention and veracity score fusion).
First, NVKE improves performance. Without it, performance drops by $0.0351$ (resp.~$0.0267$) in Acc. and $0.0316$ (resp.~$0.0367$) in Macro $\vec{F}_1$ on FakeSV (resp.~FakeTT). Additional NVKE comparisons and case studies are in Appendix~\ref{APP:Frame}.
Second, all NVKE-CEI modules are essential. Removing either content-based or evidence-based LLM reduces performance on both datasets. 
Among the three LLM-component ablations, removing the evidence-based LLM causes the largest degradation,
$~0.0572$ (resp.~$0.0635$) in Acc. and $0.0529$ (resp.~$0.1012$) in Macro $\vec{F}_1$ on FakeSV (resp.~FakeTT), indicating that the evidence-based branch contributes substantially under this setup.
Finally, judge model components are essential for integrating predictions from both LLM checkers. The text feature attention and veracity score fusion filter noisy evidence and combine complementary signals. Without fine-tuning the judge model, performance drops substantially.

\subsection{Model Explainability}
To assess the explainability of NVKE-CEI, we conduct the evaluation from three perspectives. (1) We compare the quality of explanations generated by NVKE-CEI against methods capable of producing explanations: (2) We evaluate the NVKE-CEI's explainability through a qualitative case study that demonstrates how a content-grounded explanation is generated (see Appendix~\ref{APP:Qualitative}). (3) To better understand how the judge model integrates predictions from both fact-checker LLMs, we analyze the prediction distribution among the content-based LLM, evidence-based LLM, and judge model, 
to examine how their predictions relate to the final decision.

\begin{table}[!t]
\centering
\caption{Comparison of explanation quality for the NVKE-CEI and baseline methods capable of producing explanations on the FakeTT dataset.}
\resizebox{0.48\textwidth}{!}{
\begin{tabular}{cccccc}
\toprule
 & I & S & P & R & F \\
 \midrule
 CoRAG~\cite{khaliq2024ragar} & 4.27 & 4.20 & 4.33 & 4.68 & 4.47 \\
 ReAct~\cite{react} & 4.45 & 4.81 & 4.72 & 4.69 & 4.82 \\
 ExMRD~\cite{hong2025following} & 4.92 & 4.72 & 4.63 & 4.62 & 4.89 \\
 \midrule
 \textbf{NVKE-CEI} (ours) & 4.96 & 4.90 & 4.87 & 4.70 & 4.94 \\
 \bottomrule
\end{tabular}
}
\label{TAB:EXPDATA}
\end{table}

\vpara{Quality of Explainability.}
Following previous studies~\cite{hong2025following, wang2024explainable}, we use G-Eval~\cite{liu2023g}, an LLM-based reference-free evaluation, to assess explanation quality across five metrics: (1) \textit{Informativeness} (I): provides new information and context; (2) \textit{Soundness} (S): valid and logically coherent; (3) \textit{Persuasiveness} (P): convincing and well-supported; (4) \textit{Readability} (R): proper grammar and structure; (5) \textit{Fluency} (F): smooth flow with coherent ideas. Each metric uses a 5-point Likert scale (1=poorest, 5=best). Appendix~\ref{G_eval} presents the detailed settings of G-Eval.
Results on FakeTT (Table~\ref{TAB:EXPDATA}) show NVKE-CEI achieves the highest scores across all metrics. NVKE-CEI scores 4.96 in I, outperforming ExMRD by 0.04. Notably, NVKE-CEI achieves 4.90 in S and 4.87 in P, surpassing ReAct by 0.09 and 0.15, respectively. These improvements stem from NVKE-CEI's dual-perspective fact-checking that grounds explanations in both multimodal content analysis and real-world evidence. While all methods maintain comparable R and F ($>4.6$) (excluding CoRAG), NVKE-CEI's superior I, S, and P demonstrate its ability to generate comprehensive, logically coherent, and evidence-backed explanations.

\begin{table}[!t]
\centering
\caption{Prediction Distribution of the NVKE-CEI. R: Real, F: Fake.}
\resizebox{0.48\textwidth}{!}{
\begin{tabular}{cccccc}
\toprule
Evid. & Cont. & Judge & Label & \# for FakeSV & \# for FakeTT \\
\midrule
R &	R &	R &	R &	109 & 77 \\
R &	R &	R &	F &	8 &	3 \\
R &	R &	F &	R &	30 & 4 \\ 
R &	R &	F &	F &	65 & 10 \\
R &	F &	R &	R &	3 &	18 \\
R &	F &	R &	F &	3 &	1 \\
R &	F &	F &	R &	1 &	0 \\
R &	F &	F &	F &	11 & 0 \\
F &	R &	R &	R &	67 & 42 \\
F &	R &	R &	F &	9 &	7 \\
F &	R &	F & R & 18 & 5 \\
F &	R &	F &	F & 69 & 25 \\
F &	F &	R &	R &	5 & 41 \\
F &	F &	R &	F &	4 &	7 \\
F &	F &	F &	R &	5 &	13 \\
F &	F &	F &	F &	135 & 46 \\
\bottomrule
\end{tabular}
\label{TAB:Dis}
}
\end{table}

\vpara{Prediction Distribution of the evidence-based LLM, content-based LLM, and judge model.}
To better understand how the predictions of the judge model are formed based on the content-based and evidence-based LLMs, we present the prediction distribution in Table~\ref{TAB:Dis}.
Through empirical analysis, we find that the judge model prefers to trust evidence-based LLM when strong evidence exists during conflicts, while defaulting to content-based LLM when evidence is insufficient or contradictory.

\section{Related Work}
This section reviews prior work on FNVDE and distinguishes it from our approach. Additional work on text-image fake news detection, and explainable fake news detection is reviewed in Appendix~\ref{MoreRW}.

\vpara{FNVDE.}
Detecting fake news in videos is more challenging than in text-image content because videos contain richer and more diverse modalities, making it difficult to understand and verify the information they convey~\cite{baseline1}.
Early fake news video detection studies~\cite{shang2021multimodal, li2022cnn, choi2021using, liu2023covid} primarily focused on feature extraction and integration across multiple modalities.
Subsequent studies incorporate additional context, including social information~\cite{baseline1} and neighboring videos from the same event~\cite{qi2023two}, or model the creative process of fake news videos~\cite{baseline2}. This line of work broadens the available information beyond earlier content- or metadata-focused feature pipelines~\cite{hou2019towards, palod2019misleading}.
With the pretrained knowledge and generative abilities of LLMs, recent studies~\cite{hong2025following, zhang2025fact, niu2025pioneering} have begun applying LLMs to the problem of FNVDE. 
These methods compress news videos into limited visual inputs, including sampled frames, extracted keyframes, or derived visual descriptions. Such representations may omit briefly occurring cues or temporal dependencies needed for understanding the video.
\citet{hong2025following} propose ExMRD, which leverages LLM's pretrained knowledge to detect fake news videos, while \citet{zhang2025fact} fine-tuning LLMs on video-text pairs for fake news video detection. However, these methods struggle when news videos require real-world evidence for verification, as they rely solely on the internal knowledge of LLMs. 

\section{Conclusion}

We study the FNVDE task and identify two limitations in the existing literature. 
First, commonly used frame-sampling or clustering-based selection strategies may omit veracity-relevant cues or provide insufficient temporal context for news-video understanding. 
Second, prior methods neglect either multimodal understanding or evidence retrieval. 
To address these challenges, we propose NVKE-CEI, which combines a news video keyframes extraction method with an FNVDE framework that integrates content and evidence information.
Experiments on two datasets with thirteen baselines show that the NVKE-CEI achieves SOTA detection performance while providing content-grounded explanations.

\section*{Limitations}
We discuss four main limitations of our work.
First, gradually evolving, veracity-relevant cues may not be fully captured by NVKE's similarity-based approach. NVKE selects frames around relatively large changes in combined visual and OCR-text similarity, which may miss cues that evolve gradually while maintaining high frame-to-frame similarity.
Future versions will explore temporal modeling techniques and fine-grained manipulation detection methods to capture progressive content distortions.
Second, as illustrated by the error cases in Appendix~\ref{Error}, NVKE-CEI can fail when the content-based and evidence-based LLMs converge on the same incorrect conclusion. In these cases, their agreement may leave the judge model without a sufficiently strong contradictory signal to correct the error.
Future work will explore incorporating external validation mechanisms or ensemble methods with diverse reasoning strategies to provide additional corrective signals in such cases.
Third, NVKE-CEI relies on proprietary GPT-4o APIs and live web search, so its latency, cost, retrieved evidence, and outputs may vary with service versions and availability. Under our experimental setup, processing one batch of 15 samples takes 29.6 seconds, which may hinder practical deployment.
Future work will explore adaptive routing to reduce unnecessary LLM calls and evaluations with open-source backbones.
Finally, explanation quality is evaluated only on FakeTT using a single LLM evaluator and without human judgments. Therefore, the reported G-Eval scores may reflect evaluator-specific preferences and do not establish whether the generated explanations are faithful to the final judge prediction.

\section*{Acknowledgments}
This work was supported by the Early Career Scheme (ECS) from the Research Grants Council of HKSAR (HKBU 22202423), the General Research Fund (GRF) from the Research Grants Council of HKSAR (HKBU 12203425), a grant from the Germany/Hong Kong Joint Research Scheme sponsored by the Research Grants Council of HKSAR and the German Academic Exchange Service of Germany (No. G-HKBU208/25), the Initiation Grant for Faculty Niche Research Areas 2023/24 (No. RC-FNRA-IG/23-24/COMM/01), Research Cluster Matching Scheme (No. RCMS/24-25/01) of Hong Kong Baptist University, Guangdong and Hong Kong Universities ``1+1+1'' Joint Research Collaboration Scheme (Project No. 2025A0505000001), National Natural Science Foundation of China (No. 62202402, and No. 61906161), and Startup Grant (Tier 1) for New Academics AY2020/21 of Hong Kong Baptist University.

\bibliography{main}

@inproceedings{cuconasu2024power,
  title={The power of noise: Redefining retrieval for rag systems},
  author={Cuconasu, Florin and Trappolini, Giovanni and Siciliano, Federico and Filice, Simone and Campagnano, Cesare and Maarek, Yoelle and Tonellotto, Nicola and Silvestri, Fabrizio},
  booktitle={Proc.~of ACM SIGIR},
  year={2024}
}

@inproceedings{chen2024benchmarking,
  title={Benchmarking large language models in retrieval-augmented generation},
  author={Chen, Jiawei and Lin, Hongyu and Han, Xianpei and Sun, Le},
  booktitle={Proc.~of AAAI},
  year={2024}
}

@article{xiang2024certifiably,
  title={Certifiably Robust RAG against Retrieval Corruption},
  author={Xiang, Chong and Wu, Tong and Zhong, Zexuan and Wagner, David and Chen, Danqi and Mittal, Prateek},
  journal={arXiv preprint arXiv:2405.15556},
  year={2024}
}

@inproceedings{zhou2024streaming,
  title={Streaming dense video captioning},
  author={Zhou, Xingyi and Arnab, Anurag and Buch, Shyamal and Yan, Shen and Myers, Austin and Xiong, Xuehan and Nagrani, Arsha and Schmid, Cordelia},
  booktitle={Proc.~of IEEE CVPR},
  year={2024}
}

@inproceedings{jin2024chat,
  title={Chat-univi: Unified visual representation empowers large language models with image and video understanding},
  author={Jin, Peng and Takanobu, Ryuichi and Zhang, Wancai and Cao, Xiaochun and Yuan, Li},
  booktitle={Proc.~of IEEE CVPR},
  year={2024}
}

@inproceedings{KNN,
  title={KNN model-based approach in classification},
  author={Guo, Gongde and Wang, Hui and Bell, David and Bi, Yaxin and Greer, Kieran},
  booktitle={Proc.~of OTM},
  year={2003}
}

@article{wang2021seeing,
  title={Seeing is believing? How including a video in fake news influences users’ reporting of the fake news to social media platforms},
  author={Wang, Shuting Ada and Pang, Min-Seok and Pavlou, Paul A},
  journal={MIS Quarterly},
  year={2021}
}

@article{sundar2021seeing,
  title={Seeing is believing: Is video modality more powerful in spreading fake news via online messaging apps?},
  author={Sundar, S Shyam and Molina, Maria D and Cho, Eugene},
  journal={Journal of Computer-Mediated Communication},
  volume={26},
  number={6},
  pages={301--319},
  year={2021}
}

@inproceedings{videosurvey,
  title={Combating online misinformation videos: Characterization, detection, and future directions},
  author={Bu, Yuyan and Sheng, Qiang and Cao, Juan and Qi, Peng and Wang, Danding and Li, Jintao},
  booktitle={Proc.~of ACM MM},
  year={2023}
  }

@inproceedings{baseline1,
  title={Fakesv: A multimodal benchmark with rich social context for fake news detection on short video platforms},
  author={Qi, Peng and Bu, Yuyan and Cao, Juan and Ji, Wei and Shui, Ruihao and Xiao, Junbin and Wang, Danding and Chua, Tat-Seng},
  booktitle={Proc.~of AAAI},
  year={2023}
}

@inproceedings{baseline2,
  title={FakingRecipe: Detecting Fake News on Short Video Platforms from the Perspective of Creative Process},
  author={Bu, Yuyan and Sheng, Qiang and Cao, Juan and Qi, Peng and Wang, Danding and Li, Jintao},
  booktitle={Proc.~of ACM MM},
  year={2024}
}

@inproceedings{qi2023two,
  title={Two heads are better than one: Improving fake news video detection by correlating with neighbors},
  author={Qi, Peng and Zhao, Yuyang and Shen, Yufeng and Ji, Wei and Cao, Juan and Chua, Tat-Seng},
  booktitle={Proc.~of ACL Findings},
  year={2023}
}

@article{danger1,
  title={The spread of true and false news online},
  author={Vosoughi, Soroush and Roy, Deb and Aral, Sinan},
  journal={Science},
  volume={359},
  number={6380},
  pages={1146--1151},
  year={2018},
}

@article{wang2024mfc,
  title={MFC-Bench: Benchmarking Multimodal Fact-Checking with Large Vision-Language Models},
  author={Wang, Shengkang and Lin, Hongzhan and Luo, Ziyang and Ye, Zhen and Chen, Guang and Ma, Jing},
  journal={arXiv preprint arXiv:2406.11288},
  year={2024}
}

@inproceedings{choi2021using,
  title={Using topic modeling and adversarial neural networks for fake news video detection},
  author={Choi, Hyewon and Ko, Youngjoong},
  booktitle={Proc.~of ACM CIKM},
  year={2021}
}

@inproceedings{shang2021multimodal,
  title={A multimodal misinformation detector for covid-19 short videos on tiktok},
  author={Shang, Lanyu and Kou, Ziyi and Zhang, Yang and Wang, Dong},
  booktitle={Proc.~of IEEE Big Data},
  year={2021}
}

@inproceedings{wu2024interpretable,
  title={Interpretable Short Video Rumor Detection Based on Modality Tampering},
  author={Wu, Kaixuan and Lin, Yanghao and Cao, Donglin and Lin, Dazhen},
  booktitle={Proc.~of LREC-COLING},
  year={2024}
}

@article{cheng2021socially, 
  title={Socially responsible ai algorithms: Issues, purposes, and challenges},
  author={Cheng, Lu and Varshney, Kush R and Liu, Huan},
  journal={Journal of Artificial Intelligence Research},
  volume={71},
  pages={1137--1181},
  year={2021}
}

@inproceedings{wang2023explainable,
  title={Explainable claim verification via knowledge-grounded reasoning with large language models},
  author={Wang, Haoran and Shu, Kai},
  booktitle={Proc.~of EMNLP Findings},
  year={2023}
}

@inproceedings{wang2024explainable,
  title={Explainable Fake News Detection With Large Language Model via Defense Among Competing Wisdom},
  author={Wang, Bo and Ma, Jing and Lin, Hongzhan and Yang, Zhiwei and Yang, Ruichao and Tian, Yuan and Chang, Yi},
  booktitle={Proc.~of ACM WWW},
  year={2024}
}

@inproceedings{choi2024fact,
  title={{FACT-GPT}: Fact-Checking Augmentation via Claim Matching with {LLMs}},
  author={Choi, Eun Cheol and Ferrara, Emilio},
  booktitle={Proc.~of ACM WWW short papers},
  year={2024}
}

@inproceedings{khaliq2024ragar,
  title={{RAGAR}, Your Falsehood Radar: {RAG}-Augmented Reasoning for Political Fact-Checking using Multimodal Large Language Models},
  author={Khaliq, Mohammed Abdul and Chang, Paul Yu-Chun and Ma, Mingyang and Pflugfelder, Bernhard and Mileti{\'c}, Filip},
  booktitle={Proc.~of FEVER},
  year={2024}
}

@article{gong2023multimodal,
  title={Multimodal-gpt: A vision and language model for dialogue with humans},
  author={Gong, Tao and Lyu, Chengqi and Zhang, Shilong and Wang, Yudong and Zheng, Miao and Zhao, Qian and Liu, Kuikun and Zhang, Wenwei and Luo, Ping and Chen, Kai},
  journal={arXiv preprint arXiv:2305.04790},
  year={2023}
}

@article{achiam2023gpt,
  title={{GPT-4} Technical Report},
  author={{OpenAI}},
  journal={arXiv preprint arXiv:2303.08774},
  year={2023}
}

@inproceedings{ji2023towards,
  title={Towards mitigating LLM hallucination via self reflection},
  author={Ji, Ziwei and Yu, Tiezheng and Xu, Yan and Lee, Nayeon and Ishii, Etsuko and Fung, Pascale},
  booktitle={Proc.~of EMNLP Findings},
  year={2023}
}

@inproceedings{perkovic2024hallucinations,
  title={Hallucinations in llms: Understanding and addressing challenges},
  author={Perkovi{\'c}, Gabrijela and Drobnjak, Antun and Boti{\v{c}}ki, Ivica},
  booktitle={Proc.~of IEEE MIPRO},
  year={2024}
}

@inproceedings{hu2024bad,
  title={Bad actor, good advisor: Exploring the role of large language models in fake news detection},
  author={Hu, Beizhe and Sheng, Qiang and Cao, Juan and Shi, Yuhui and Li, Yang and Wang, Danding and Qi, Peng},
  booktitle={Proc.~of AAAI},
  year={2024}
}

@inproceedings{asai2023self,
  title={Self-rag: Learning to retrieve, generate, and critique through self-reflection},
  author={Asai, Akari and Wu, Zeqiu and Wang, Yizhong and Sil, Avirup and Hajishirzi, Hannaneh},
  booktitle={Proc.~of ICLR},
  year={2024}
}

@inproceedings{welz2024enhancing,
  title={Enhancing Large Language Models Through External Domain Knowledge},
  author={Welz, Laslo and Lanquillon, Carsten},
  booktitle={Proc.~of AI-HCI},
  year={2024}
}

@inproceedings{lewis2020retrieval,
  title={Retrieval-Augmented Generation for Knowledge-Intensive {NLP} Tasks},
  author={Lewis, Patrick and Perez, Ethan and Piktus, Aleksandra and Petroni, Fabio and Karpukhin, Vladimir and Goyal, Naman and K{\"u}ttler, Heinrich and Lewis, Mike and Yih, Wen-tau and Rockt{\"a}schel, Tim and Riedel, Sebastian and Kiela, Douwe},
  booktitle={Proc.~of NeurIPS},
  year={2020}
}

@inproceedings{borgeaud2022improving,
  title={Improving Language Models by Retrieving from Trillions of Tokens},
  author={Borgeaud, Sebastian and Mensch, Arthur and Hoffmann, Jordan and Cai, Trevor and Rutherford, Eliza and Millican, Katie and van den Driessche, George and Lespiau, Jean-Baptiste and Damoc, Bogdan and Clark, Aidan and de Las Casas, Diego and Guy, Aurelia and Menick, Jacob and Ring, Roman and Hennigan, Tom and Huang, Saffron and Maggiore, Loren and Jones, Chris and Cassirer, Albin and Brock, Andy and Paganini, Michela and Irving, Geoffrey and Vinyals, Oriol and Osindero, Simon and Simonyan, Karen and Rae, Jack W. and Elsen, Erich and Sifre, Laurent},
  booktitle={Proc.~of ICML},
  year={2022}
}

@inproceedings{radford2021learning,
  title={Learning Transferable Visual Models from Natural Language Supervision},
  author={Radford, Alec and Kim, Jong Wook and Hallacy, Chris and Ramesh, Aditya and Goh, Gabriel and Agarwal, Sandhini and Sastry, Girish and Askell, Amanda and Mishkin, Pamela and Clark, Jack and Krueger, Gretchen and Sutskever, Ilya},
  booktitle={Proc.~of ICML},
  year={2021}
}

@article{cao2020exploring,
  title={Exploring the role of visual content in fake news detection},
  author={Cao, Juan and Qi, Peng and Sheng, Qiang and Yang, Tianyun and Guo, Junbo and Li, Jintao},
  journal={Disinformation, Misinformation, and Fake News in Social Media: Emerging Research Challenges and Opportunities},
  pages={141--161},
  year={2020}
}

@article{tufchi2023comprehensive,
  title={A comprehensive survey of multimodal fake news detection techniques: advances, challenges, and opportunities},
  author={Tufchi, Shivani and Yadav, Ashima and Ahmed, Tanveer},
  journal={International Journal of Multimedia Information Retrieval},
  volume={12},
  number={2},
  pages={28},
  year={2023}
}

@inproceedings{alam2021survey,
  title={A survey on multimodal disinformation detection},
  author={Alam, Firoj and Cresci, Stefano and Chakraborty, Tanmoy and Silvestri, Fabrizio and Dimitrov, Dimiter and Martino, Giovanni Da San and Shaar, Shaden and Firooz, Hamed and Nakov, Preslav},
  booktitle={Proc.~of COLING},
  year={2022}
}

@inproceedings{jin2017multimodal,
  title={Multimodal fusion with recurrent neural networks for rumor detection on microblogs},
  author={Jin, Zhiwei and Cao, Juan and Guo, Han and Zhang, Yongdong and Luo, Jiebo},
  booktitle={Proc.~of ACM MM},
  year={2017}
}

@inproceedings{khattar2019mvae,
  title={Mvae: Multimodal variational autoencoder for fake news detection},
  author={Khattar, Dhruv and Goud, Jaipal Singh and Gupta, Manish and Varma, Vasudeva},
  booktitle={Proc.~of ACM WWW},
  year={2019}
}

@inproceedings{chen2022cross,
  title={Cross-modal ambiguity learning for multimodal fake news detection},
  author={Chen, Yixuan and Li, Dongsheng and Zhang, Peng and Sui, Jie and Lv, Qin and Tun, Lu and Shang, Li},
  booktitle={Proc.~of ACM WWW},
  year={2022}
}

@inproceedings{hu2023large,
  title={Do Large Language Models Know about Facts?},
  author={Hu, Xuming and Chen, Junzhe and Li, Xiaochuan and Guo, Yufei and Wen, Lijie and Yu, Philip S and Guo, Zhijiang},
  booktitle={Proc.~of ICLR},
  year={2024}
}

@inproceedings{chen2023can,
  title={Can llm-generated misinformation be detected?},
  author={Chen, Canyu and Shu, Kai},
  booktitle={Proc.~of ICLR},
  year={2024}
}

@inproceedings{liu2025detect,
  title={Detect, investigate, judge and determine: A knowledge-guided framework for few-shot fake news detection},
  author={Liu, Ye and Zhu, Jiajun and Liu, Xukai and Tang, Haoyu and Zhang, Yanghai and Zhang, Kai and Zhou, Xiaofang and Chen, Enhong},
  booktitle={Proc.~of IEEE ICDM},
  year={2025}
}

@inproceedings{singhal2024evidence,
  title={Evidence-backed fact checking using RAG and few-shot in-context learning with LLMs},
  author={Singal, Ronit and Patwa, Pransh and Patwa, Parth and Chadha, Aman and Das, Amitava},
  booktitle={Proc.~of FEVER},
  year={2024}
}

@inproceedings{hou2019towards,
  title={Towards automatic detection of misinformation in online medical videos},
  author={Hou, Rui and P{\'e}rez-Rosas, Ver{\'o}nica and Loeb, Stacy and Mihalcea, Rada},
  booktitle={Proc.~of ICMI},
  year={2019}
}

@inproceedings{palod2019misleading,
  title={Misleading metadata detection on YouTube},
  author={Palod, Priyank and Patwari, Ayush and Bahety, Sudhanshu and Bagchi, Saurabh and Goyal, Pawan},
  booktitle={Proc.~of ECIR},
  year={2019}
}

@article{li2022cnn,
  title={A CNN-based misleading video detection model},
  author={Li, Xiaojun and Xiao, Xvhao and Li, Jia and Hu, Changhua and Yao, Junping and Li, Shaochen},
  journal={Scientific Reports},
  volume={12},
  number={1},
  pages={6092},
  year={2022}
}

@inproceedings{liu2023covid,
  title={Covid-vts: Fact extraction and verification on short video platforms},
  author={Liu, Fuxiao and Yacoob, Yaser and Shrivastava, Abhinav},
  booktitle={Proc.~of EACL},
  year={2023}
}

@article{guo2022survey,
  title={A survey on automated fact-checking},
  author={Guo, Zhijiang and Schlichtkrull, Michael and Vlachos, Andreas},
  journal={Transactions of the Association for Computational Linguistics},
  volume={10},
  pages={178--206},
  year={2022}
}

@inproceedings{vlachos2014fact,
  title={Fact checking: Task definition and dataset construction},
  author={Vlachos, Andreas and Riedel, Sebastian},
  booktitle={Proc.~of the ACL Workshop on Language Technologies and Computational Social Science},
  year={2014}
}

@article{brandtzaeg2018journalists,
  title={How journalists and social media users perceive online fact-checking and verification services},
  author={Brandtzaeg, Petter Bae and F{\o}lstad, Asbj{\o}rn and Chaparro Dom{\'\i}nguez, Mar{\'\i}a {\'A}ngeles},
  journal={Journalism Practice},
  volume={12},
  number={9},
  pages={1109--1129},
  year={2018}
}

@inproceedings{react,
  title={React: Synergizing reasoning and acting in language models},
  author={Yao, Shunyu and Zhao, Jeffrey and Yu, Dian and Du, Nan and Shafran, Izhak and Narasimhan, Karthik and Cao, Yuan},
  booktitle={Proc.~of ICLR},
  year={2023}
}

@article{chen2025multimodal,
  title={Multimodal Fake News Video Explanation: Dataset, Model and Evaluation},
  author={Lizhi Chen and Zhong Qian and Peifeng Li and Qiaoming Zhu},
  journal={Information Processing \& Management},
  volume={63},
  number={6},
  pages={104769},
  year={2026},
}

@inproceedings{luo2024message,
  title={Message Injection Attack on Rumor Detection under the Black-Box Evasion Setting Using Large Language Model},
  author={Luo, Yifeng and Li, Yupeng and Wen, Dacheng and Lan, Liang},
  booktitle={Proc.~of ACM WWW},
  year={2024}
}

@inproceedings{luo2026cheap,
  title={From Cheap Fakes to Pure Synthesis: Addressing the New Era of T2V Fake News Videos},
  author={Luo, Yifeng and Li, Yupeng and Lan, Liang and Wang, Tian},
  booktitle={Proc.~of ACM MM},
  year={2026}
}

@inproceedings{he2026novel,
  title={Novel Claim or Déjà Vu? Rethinking ``Contamination-Free'' Dynamic Evaluation for Multimodal Automated Fact-Checking}, 
  author={He, Haorui and Chen, Xinwen and Wen, Dacheng and Cheng, Reynold and Lau, Francis C. M. and Li, Yupeng},
  booktitle={Proc.~of ACM MM},
  year={2026},
}

@inproceedings{he2026fact2fiction,
  title={Fact2Fiction: Targeted poisoning attack to agentic fact-checking system},
  author={He, Haorui and Li, Yupeng and Zhu, Bin Benjamin and Wen, Dacheng and Cheng, Reynold and Lau, Francis C. M.},
  booktitle={Proc.~of AAAI},
  year={2026}
}

@inproceedings{he2026debating,
  title={Debating truth: Debate-driven claim verification with multiple large language model agents},
  author={He, Haorui and Li, Yupeng and Wen, Dacheng and Chen, Yang and Cheng, Reynold and Chen, Donglong and Lau, Francis C. M.},
  booktitle={Proc.~of ACM WWW},
  year={2026}
}

@inproceedings{li2024mcfend,
  title={MCFEND: a multi-source benchmark dataset for Chinese fake news detection},
  author={Li, Yupeng and He, Haorui and Bai, Jin and Wen, Dacheng},
  booktitle={Proc.~of ACM WWW},
  year={2024}
}

@inproceedings{hong2025following,
  title={Following clues, approaching the truth: Explainable micro-video rumor detection via chain-of-thought reasoning},
  author={Hong, Rongpei and Lang, Jian and Xu, Jin and Cheng, Zhangtao and Zhong, Ting and Zhou, Fan},
  booktitle={Proc.~of ACM WWW},
  year={2025}
}

@article{deepseek,
  title={{DeepSeek-V3} Technical Report},
  author={{DeepSeek-AI}},
  journal={arXiv preprint arXiv:2412.19437},
  year={2024}
}

@article{qwen,
  title={{Qwen2-VL}: Enhancing Vision-Language Model's Perception of the World at Any Resolution},
  author={Wang, Peng and Bai, Shuai and Tan, Sinan and Wang, Shijie and Fan, Zhihao and Bai, Jinze and Chen, Keqin and Liu, Xuejing and Wang, Jialin and Ge, Wenbin and Fan, Yang and Dang, Kai and Du, Mengfei and Ren, Xuancheng and Men, Rui and Liu, Dayiheng and Zhou, Chang and Zhou, Jingren and Lin, Junyang},
  journal={arXiv preprint arXiv:2409.12191},
  year={2024}
}

@inproceedings{niu2025pioneering,
  title={Pioneering Explainable Video Fact-Checking with a New Dataset and Multi-role Multimodal Model Approach},
  author={Niu, Kaipeng and Xu, Danni and Yang, Bingjian and Liu, Wenxuan and Wang, Zheng},
  booktitle={Proc.~of AAAI},
  year={2025}
}

@inproceedings{zhang2025fact,
  title={Fact-R1: Towards Explainable Video Misinformation Detection with Deep Reasoning},
  author={Zhang, Fanrui and Li, Dian and Zhang, Qiang and Lin, Junxiong and Yan, Jiahong and Liu, Jiawei and Zha, Zheng-Jun},
  booktitle={Proc.~of NeurIPS},
  year={2025}
}

@article{yan2025debunk,
  title={Debunk and Infer: Multimodal Fake News Detection via Diffusion-Generated Evidence and LLM Reasoning},
  author={Yan, Kaiying and Liu, Moyang and Liu, Yukun and Fu, Ruibo and Wen, Zhengqi and Tao, Jianhua and Liu, Xuefei},
  journal={arXiv preprint arXiv:2506.21557},
  year={2025}
}

@inproceedings{wang2024rolellm,
  title={{RoleLLM}: Benchmarking, Eliciting, and Enhancing Role-Playing Abilities of Large Language Models},
  author={Wang, Noah and Peng, Z.Y. and Que, Haoran and Liu, Jiaheng and Zhou, Wangchunshu and Wu, Yuhan and Guo, Hongcheng and Gan, Ruitong and Ni, Zehao and Yang, Jian and Zhang, Man and Zhang, Zhaoxiang and Ouyang, Wanli and Xu, Ke and Huang, Wenhao and Fu, Jie and Peng, Junran},
  booktitle={Proc.~of ACL Findings},
  year={2024}
}

@article{shanahan2023role,
  title={Role play with large language models},
  author={Shanahan, Murray and McDonell, Kyle and Reynolds, Laria},
  journal={Nature},
  volume={623},
  number={7987},
  pages={493--498},
  year={2023}
}

@inproceedings{liu2023g,
  title={G-eval: NLG evaluation using gpt-4 with better human alignment},
  author={Liu, Yang and Iter, Dan and Xu, Yichong and Wang, Shuohang and Xu, Ruochen and Zhu, Chenguang},
  booktitle={Proc.~of EMNLP},
  year={2023}
}

@article{team2025tongyi,
  title={Tongyi {DeepResearch} Technical Report},
  author={{Tongyi DeepResearch Team}},
  journal={arXiv preprint arXiv:2510.24701},
  year={2025}
}

@inproceedings{ge2025resolving,
author = {Ge, Ziyu and Wu, Yuhao and Chin, Daniel Wai Kit and Lee, Roy Ka-Wei and Cao, Rui},
title = {Resolving conflicting evidence in automated fact-checking: a study on retrieval-augmented LLMs},
booktitle = {Proc.~of IJCAI},
year = {2025}
}

@inproceedings{cheng2024dated,
  title={Dated data: Tracing knowledge cutoffs in large language models},
  author={Cheng, Jeffrey and Marone, Marc and Weller, Orion and Lawrie, Dawn and Khashabi, Daniel and Van Durme, Benjamin},
  booktitle={Proc.~of COLM},
  year={2024}
}

@inproceedings{wan2025unveiling,
  title={Unveiling confirmation bias in chain-of-thought reasoning},
  author={Wan, Yue and Jia, Xiaowei and Li, Xiang Lorraine},
  booktitle={Proc.~of ACL Findings},
  year={2025}
}

@inproceedings{wang2025truth,
  title={When truth is overridden: Uncovering the internal origins of sycophancy in large language models},
  author={Wang, Keyu and Li, Jin and Yang, Shu and Zhang, Zhuoran and Wang, Di},
  booktitle={Proc.~of AAAI},
  year={2026}
}

\appendix
\nobalance

\section{Pilot Study on FakeSV Dataset}
\label{Pilot}

\begin{table*}[!t]
\centering
\caption{Class-conditional prevalence of frame-selection challenge labels in the analyzed FakeSV videos.}
\resizebox{0.7\textwidth}{!}{
\begin{tabular}{cccc}
\toprule
Problem Type & Fake News Video Ratio & Real News Video Ratio & Difference \\
\midrule
Semantic redundancy & 82.7\% & 50.8\% & Fake +31.9\%\\
Keyframe sparsity & 19.1\% & 25.1\% & Real +6.0\% \\
Cross-frame reasoning gap & 3.2\% & 9.5\% & Real +6.3\% \\
Temporal discontinuity & 8.1\% & 9.0\% & Equivalent \\
\bottomrule
\end{tabular}
\label{TAB:ProblemFakeSV}
}
\end{table*}

As elaborated in the introduction, effective video frame selection is critical for FNVDE. Existing LLM-based methods mostly use random frame sampling, while some video understanding methods adopt density-based keyframe extraction. 
These strategies may omit briefly occurring but veracity-relevant cues or provide insufficient temporal context for understanding the news video.
To characterize these potential frame-selection challenges and motivate our proposed News Video Keyframes Extraction (NVKE) method, we conduct a pilot study on the FakeSV dataset.
The core goal is to estimate the prevalence of these challenges and analyze their distributions across real and fake news videos.
Specifically, we employ GPT-4o-mini as the LLM for this study, with the following experimental procedure: (1) Sample one frame per second from each news video, then stitch every 4 consecutive frames into a single image; (2) Input all stitched images, the text material of the news video, and the veracity label (real/fake) of the news video into the LLM; (3) Define four types of sampling problems for the LLM to identify, including semantic redundancy (videos contain a large amount of static frames), keyframe sparsity (key evidence may appear in only a few frames), cross-frame reasoning gap (veracity assessment requires integrating information from multiple frames), and temporal discontinuity (contradictions require comparing consecutive frames). 
The LLM assigns any applicable challenge labels; if none applies, it returns the separate \textit{None of the Above} label.
To reduce decoding variability in our pilot study, we set the temperature to 0.
The prompt for our pilot study can be found in Appendix~\ref{P:Pilot}.

We validate this stability by repeating the pilot study 5 times, observing negligible variance in the results with mean $93.4\%$ and standard deviation $0.12\%$. This confirms that the identified sampling issues are consistent characteristics of the dataset rather than artifacts of model randomness.
Table~\ref{TAB:ProblemFakeSV} reports the results from the first run.
The LLM assigned at least one of the four frame-selection challenge labels to $93.4\%$ of the analyzed videos, indicating that these challenges are common in FakeSV.
Specifically, semantic redundancy is the most prominent problem (2,419 videos, 66.7\%), followed by keyframe sparsity (801 videos, 22.1\%), temporal discontinuity (311 videos, 8.6\%), and cross-frame reasoning gap (230 videos, 6.3\%). There are descriptive differences between real and fake news: fake news has 31.9\% higher semantic redundancy (82.7\% vs 50.8\%), while real news has higher keyframe sparsity (25.1\% vs 19.1\%) and cross-frame reasoning gap (9.5\% vs 3.2\%), with equivalent temporal discontinuity (8.1\% vs 9.0\%). 
Statistically, the top two common label combinations are single semantic redundancy (2,267 videos, 62.6\%) and single keyframe sparsity (546 videos, 15.1\%); on average, each video has an average of 1.04 problems, and the most common two-label combination is semantic redundancy with cross-frame reasoning gap (50 videos, 1.4\%).
In summary, semantic redundancy is more frequently identified in fake news videos, whereas keyframe sparsity and cross-frame reasoning gap are more frequently identified in real news videos.

\begin{table*}[!t]
\centering
\caption{More experiments on different LLM backbones.}
\resizebox{0.9\textwidth}{!}{
\begin{tabular}{ccccccccccc}
\toprule
\multicolumn{2}{c}{} & \multicolumn{4}{c}{\textbf{FakeSV} \cite{baseline1}} &  & \multicolumn{4}{c}{\textbf{FakeTT} \cite{baseline2}} \\
\cmidrule{3-6} \cmidrule{8-11}
Method & Class & Acc. & Macro $\vec{F}_1$ & Prec. & Rec. &  & Acc. & Macro $\vec{F}_1$ & Prec. & Rec. \\
\midrule
\multirow{2}{*}{NVKE-CEI + Qwen3-vl-plus} & Fake & \multirow{2}{*}{0.8506} & \multirow{2}{*}{0.8493} & 0.8885 & 0.8388 & & \multirow{2}{*}{0.8428} & \multirow{2}{*}{0.8300} & 0.7203 & 0.8586 \\
& Real & & & 0.8078 & 0.8655 & & & & 0.9227 & 0.8350 \\
\multirow{2}{*}{NVKE-CEI + GPT-4o} & Fake & \multirow{2}{*}{0.8561} & \multirow{2}{*}{0.8514} & 0.8383 & 0.9211 & & \multirow{2}{*}{0.8662} & \multirow{2}{*}{0.8505} & 0.7864 & 0.8182 \\
& Real & & & 0.8846 & 0.7731 & & & & 0.9082 & 0.8900 \\
\bottomrule
\end{tabular}
\label{TAB:Backbone}
}
\end{table*}

\begin{table*}[!t]
\centering
\caption{Comparison of NVKE with density-based keyframe extraction.}
\resizebox{0.9\textwidth}{!}{
\begin{tabular}{ccccccccccc}
\toprule
\multicolumn{2}{c}{} & \multicolumn{4}{c}{\textbf{FakeSV} \cite{baseline1}} &  & \multicolumn{4}{c}{\textbf{FakeTT} \cite{baseline2}} \\
\cmidrule{3-6} \cmidrule{8-11}
Method & Class & Acc. & Macro $\vec{F}_1$ & Prec. & Rec. &  & Acc. & Macro $\vec{F}_1$ & Prec. & Rec. \\
\midrule
\multirow{2}{*}{\textbf{ReAct + KNN}~\cite{react, KNN}} & Fake & \multirow{2}{*}{0.7711} & \multirow{2}{*}{0.7709} & 0.8744 & 0.7013 & & \multirow{2}{*}{0.7419} & \multirow{2}{*}{0.7306} & 0.5905 & 0.7894 \\
& Real & & & 0.6832 & 0.8648 & & & & 0.8684 & 0.7173 \\
\multirow{2}{*}{\textbf{NVKE-CEI} (ours)} & Fake & \multirow{2}{*}{0.8561} & \multirow{2}{*}{0.8514} & 0.8383 & 0.9211 & & \multirow{2}{*}{0.8662} & \multirow{2}{*}{0.8505} & 0.7864 & 0.8182 \\
& Real & & & 0.8846 & 0.7731 & & & & 0.9082 & 0.8900 \\
\bottomrule
\end{tabular}
\label{TAB:Qw}
}
\end{table*}

\section{More Details of the Experiment Setting}
\label{EXP:MOREEXP}

\subsection{More Details of the Datasets and Evaluation Metrics}
\label{APP:dataset}
In our experiments, we consider two widely used datasets in the literature on the detection of fake news videos, FakeSV~\cite{baseline1} and FakeTT~\cite{baseline2}. FakeSV~\cite{baseline1} is currently the largest Chinese dataset for detecting fake news videos and collects its data from two popular Chinese short video platforms, i.e., \textit{Douyin} and \textit{Kuaishou}. On the other hand, FakeTT~\cite{baseline2} is an English dataset sourced from the popular English-speaking short video platform, \textit{TikTok}. 
To evaluate the performance of our proposed fake news video detection method, we employ two evaluation metrics: accuracy (Acc.) and Macro $\vec{F}_1$ score. We focus primarily on the Acc. and the Macro $\vec{F}_1$ score since the Macro $\vec{F}_1$ score inherently reflects the balance between Precision (Prec.) and Recall (Rec.). 
Note that when evidence is insufficient or contradictory, zero-shot LLM baselines such as GPT-4o may output uncertain predictions (e.g., \textit{cannot determine} or \textit{failed}). Following \citet{khaliq2024ragar}, we treat all uncertain predictions as incorrect during evaluation for a fair comparison. 
This explains why the reported Prec. and Rec. values may not mathematically correspond to 
Acc. and Macro $\vec{F}_1$ scores. the Macro $\vec{F}_1$ include uncertain cases as classification errors, while the Acc. only consider definitive Real and Fake predictions.

\subsection{Baselines Implementation Details}
\label{APP:baseline}
We first introduce the implementation details of the zero-shot LLM-based baselines. For \textbf{GPT-4o}, \textbf{DeepSeek-V3}, \textbf{Qwen3-vl-plus}, and \textbf{Claude-3-5-sonnet}, we use a zero-shot prompt template based on the \citet{hu2024bad}. \textbf{ReAct} is a prompt engineering method for synergizing reasoning and acting for LLMs, where \textbf{GPT-4o} is employed for language generation. For \textbf{CoRAG}, GPT-4o is employed as the LLM to process the multimodal content and predict the veracity of the news videos. Specifically, we input the same text material used for the GPT-4o baseline and the thumbnail of the news video to GPT-4o. 
Next, we introduce the implementation details of the supervised baselines.
We follow the hyper-parameters and training-testing list from their original studies. Recall that we utilize FakeSV and FakeTT datasets in our experiments. Notably, the FakeTT dataset does not include video social context, e.g., user profiles or comment interactions. Therefore, similar to \citet{baseline2}, \textbf{SVFEND} is adapted to focus exclusively on the video content. 

\subsection{NVKE-CEI Implementation Details}
\label{APP:CEIIMP}

For the hyper-parameters in NVKE, we set $\alpha=0.5$ and $K=4$. Prior to finalizing these settings, we conducted preliminary experiments on a subset of the FakeSV dataset, i.e., 300 data samples for training and 100 data samples for testing. For $\alpha$, we selected 0.5 to balance visual and semantic information, as performance degrades when relying solely on visual features ($\alpha=1$) or OCR text ($\alpha=0$). For $K$, our experiments show that increasing from 1 to 4 consistently improves performance, but beyond 4, gains become negligible while LLM API processing time increases significantly, making $K=4$ the most cost-effective choice.
For our proposed CEI framework, we utilize GPT-4o, GPT-4o-2024-08-06, as our backbone LLM and set temperature to 0. 
To demonstrate the effectiveness of the CEI framework across different LLM backbones, we additionally conduct experiments using Qwen3-vl-plus. Details are provided in the Appendix~\ref{APP:LLMBack}.
We use GPT-4o-audio-preview to generate the audio transcripts included in the text material $M$.
For the RAG implementations (baselines and our method), we employ DuckDuckGo\footnote{\url{https://duckduckgo.com}} to retrieve real-world evidence from the Internet.
We restrict the number of retrieved results for each search query to $K_\text{RAG}=5$. We temporally restrict the RAG retriever by only collecting the information that was published before the news was fact-checked to provide the LLM with facts relevant to the time frame of the fact-checking.
For the judge model, we utilize a pre-trained CLIP model \textit{OFA-Sys/chinese-clip-vit-large-patch14} for the FakeSV dataset and \textit{openai/clip-vit-large-patch14} for the FakeTT dataset. 
We train 5 (resp.~2) epochs on FakeSV (resp.~FakeTT) datasets.

\subsection{Latency and Performance Report and Discussion}
\label{App:report}
We utilize the LangChain framework\footnote{\url{https://www.langchain.com/}} to build the content-based and evidence-based LLMs. These two LLMs run in parallel, leveraging asynchronous calls. This design allows the content-based and evidence-based LLMs to process news videos in batches simultaneously. We set the batch size to 15, meaning both LLMs handle 15 news videos per batch. On average, it takes approximately 30s for both the content-based and evidence-based LLMs to complete processing a single batch. Table~\ref{Latency} presents the time latency cost of our proposed NVKE-CEI framework compared with other LLM-based methods that utilize commercial LLM API on the FakeSV dataset. 
We report the mean wall-clock latency per 15-video batch and the overall Acc.
Our proposed NVKE-CEI achieves the highest accuracy and the second-lowest time cost. While GPT-4o has the lowest time cost, our method outperforms it in accuracy. 
Under this setup, NVKE-CEI achieves the highest accuracy and the second-lowest latency among the five evaluated API-based methods.

\begin{table}[!t]
\centering
\caption{Mean wall-clock latency per 15-video batch on FakeSV.}
\resizebox{0.3\textwidth}{!}{
\begin{tabular}{ccc}
\toprule
Method & Time Cost (s) & Acc. \\
\midrule
GPT-4o & 12.7 & 0.6518 \\
CoRAG & 121.0 & 0.3326 \\
ReAct & 29.8 & 0.6084 \\
3MFact & 34.3 & 0.8122 \\
NVKE-CEI (ours) & 29.6 & 0.8561 \\
\bottomrule
\end{tabular}
\label{Latency}
}
\end{table}

\section{More Experiments on Different LLM Backbones}
\label{APP:LLMBack}
To demonstrate the effectiveness of the NVKE-CEI framework across different LLM backbones, we additionally conduct experiments using Qwen3-vl-plus. The detailed experiment results are presented in Table~\ref{TAB:Backbone}.

\section{More Experiments on NVKE and Case Study}
\label{APP:Frame}

To better demonstrate the effectiveness of our proposed NVKE method, we compare our proposed NVKE method with the traditional density-based keyframe extraction method. The experiment results can be found in Table~\ref{TAB:Qw}. Additionally, we present a case study of our proposed NVKE-CEI framework comparing scenarios with and without frame inputs (See Table~\ref{TAB:FRAME}). Our experimental results and case analysis reveal two key findings: (1) 
NVKE-CEI outperforms the evaluated ReAct+KNN baseline, and (2) frame information is decisive in this illustrative case for effective fake news video detection.

\begin{table*}[!t]
\centering
\caption{Frame and no frame case study.}
\resizebox{0.88\textwidth}{!}{
\begin{tabular}{>{\centering\arraybackslash}cm{15mm}m{62mm}m{62mm}}
\toprule
 & & \textbf{Frame} & \textbf{No Frame} \\
\midrule
\textbf{Fake News Video} & & \multicolumn{2}{c}{\begin{minipage}[b]{0.3\columnwidth}
            \centering
            \includegraphics[width=\linewidth]{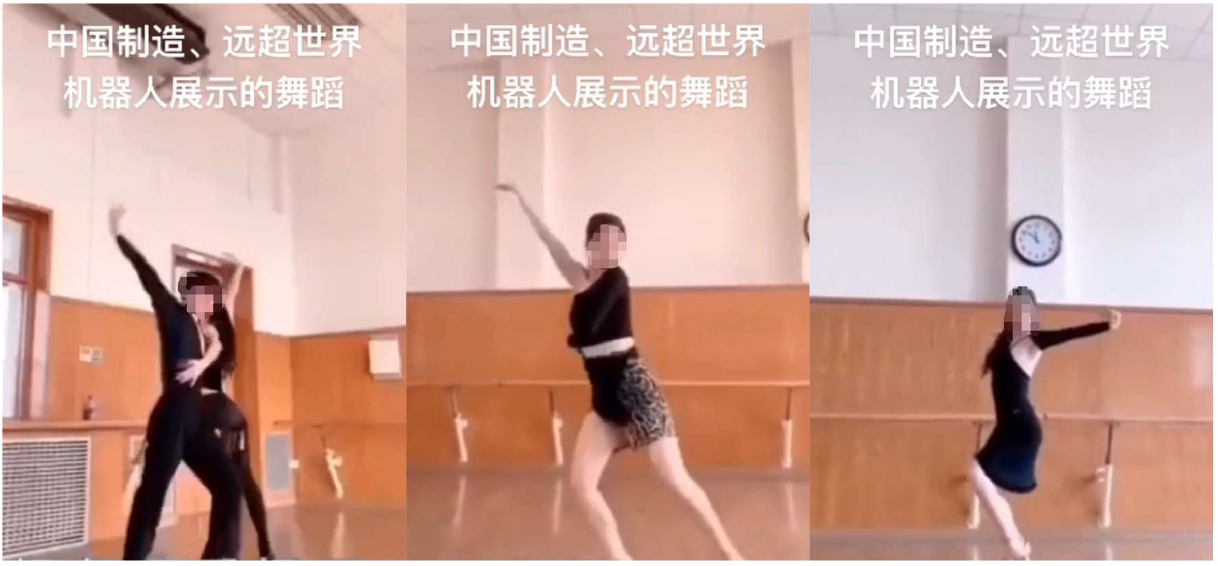}
            \end{minipage}} \\
\textbf{Corresponding Text} & & \multicolumn{2}{l}{Chinese-made robots showcased a dance that far surpassed global standards.} \\
\textbf{News Summary} & & \multicolumn{2}{m{100mm}}{The video showcases a dance performance by a domestically-produced robot, emphasizing that Chinese-made robot technology now far surpasses global standards. The ticket price is 499 yuan.} \\
\midrule
\multirow{2}{*}{\textbf{Content-based LLM}} & Reasoning Process & The video features a female dancing, claimed to be a robot performance. However, \textcolor{red}{screenshots suggest the performer is human, not robotic}. The title and OCR text stating ``robot dance performance'' may be misleading. Additionally, details like ticket prices and queue times seem intended to enhance credibility, yet no sufficient evidence proves the performer is a robot. & The video claims to display a ``Made in China, world-surpassing robot dance,'' suggesting an exhibition. However, the term ``world-surpassing'' appears exaggerated and implies a persuasive intent. OCR data shows ticket prices but does not validate the event's authenticity. Given these factors, the video might be exaggerating or misleading about the robot’s technological prowess, potentially constituting fake news.\\
 & Prediction & Fake & Fake \\
\midrule
 \multirow{10}{*}{\textbf{Evidence-based LLM}} & Query Generation & & \\
 & Reasoning Process & The video claims to show an advanced, domestically produced robot dancing. Verification is needed to confirm the robot's origin and technological capabilities as described. & The video highlights a ``Made in China'' robot exhibition, but it's essential to verify the pricing and the event's occurrence. I will research similar robot events to confirm the authenticity of this news. \\
 & Queries & ``\textit{China robot dance performance}'', ``\textit{Chinese-made robot technology}'', ``\textit{Robot dance performance tickets}'' & ``\textit{Chinese Robot Expo}'', ``\textit{Robot Dance Display}'', ``\textit{Chinese-Made Robots}'' \\
 \cmidrule{2-4}
 & Query Re-generation & & \\
 & Reasoning Process & Current search terms relate to the video but should include ticket prices and wait times for better authenticity verification. & No need to regenerate queries. \\
 & Re-generate Queries & ``\textit{Chinese Robot Dance Performance 499 Yuan Ticket}'', ``\textit{Chinese-Made Robot Dance Four-Hour Wait}'', ``\textit{Robot Dance Performance 499 Yuan Ticket}'' & None \\
 \cmidrule{2-4}
& Retrieved Evidence & \multicolumn{2}{c}{\textit{Omitted for space}} \\
 \cmidrule{2-4}
 & Veracity Prediction & & \\
 & Reasoning Process & Search results suggest the video may be misleading. Multiple sources indicate the supposed robot dance performance is actually done by humans, and the related queue and ticket information are also considered false. &  The news confirms that Chinese-made robots showcased a world-class dance display. \\
 & Prediction & Fake & Real \\
\midrule
& Final Prediction & Fake & Real \\
\midrule
\textbf{Label} & & Fake & Fake \\ 
\bottomrule
\end{tabular}
}
\label{TAB:FRAME}
\end{table*}

\begin{table}[!t]
\centering
\caption{Case study of correctly detected fake news video on the FakeTT dataset.}
\resizebox{0.369\textwidth}{!}{
\begin{tabular}{>{\centering\arraybackslash}p{22mm}m{72mm}}
\toprule
  & \textbf{FakeTT} \\
 \midrule
 \textbf{News Video} &  \begin{minipage}[b]{0.25\columnwidth}
            \centering
            \includegraphics[width=\linewidth]{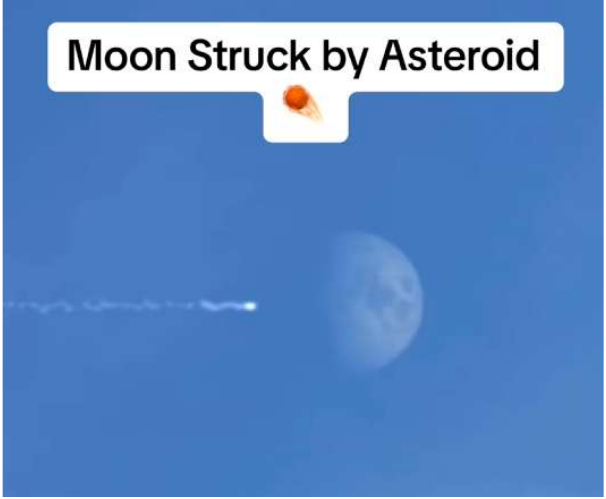}
            \end{minipage}\\
\textbf{Corresponding Text}  & Scary \#fyp \#moon \#astroid \#scary\\
\textbf{News Summary}   & The news video claims that the moon was struck by an asteroid, featuring hashtags such as \#moon \#astroid \#scary and describing the event as scary. \\
\midrule
\multicolumn{2}{l}{\textbf{Content-based LLM}} \\
Reasoning Process & The content of the video describes an event where an asteroid strikes the moon. The OCR text on the keyframes consistently repeats the phrase ``Moon Struck by Asteroid.'' This event is highly unlikely and there has been no credible news or scientific reporting on such a recent event where the moon was struck by a sizable asteroid. Such an event would be a significant piece of news globally. Furthermore, the images appear to be computer-generated or digitally manipulated, containing effects that are not commonly associated with real astronomical events. The video seems to be trying to invoke fear or shock with tags like ``\#scary'', ``\#bad'', indicating an intention to sensationalize or misinform rather than educate. Based on the elements above, the content seems to be misleading and false, fitting the criteria of a fake news video. \\
Prediction & Fake \\
\midrule
\multicolumn{2}{l}{\textbf{Evidence-based LLM}} \\
\multicolumn{2}{l}{\textit{Query Generation}} \\
Reasoning Process & The video claims that the moon has been struck by an asteroid, which would be a significant astronomical event. I need to verify if there have been any recent reports or news articles about such an event. \\
Queries & ``\textit{Moon struck by asteroid October 2023}'', ``\textit{Asteroid collision with moon news}'', ``\textit{Recent asteroid moon impact event}''\\
\multicolumn{2}{l}{\textit{Query Regeneration}} \\
Reasoning Process &  No need to regenerate queries. \\
Regenerated Queries & None \\
Retrieved Evidence & \textit{Omitted for space} \\
\multicolumn{2}{l}{\textit{Veracity Prediction}} \\
Reasoning Process & There are no credible news reports or scientific confirmations of a recent asteroid impacting the moon as claimed in the video. The searches mainly yielded results related to potential future impacts but nothing conclusive about a current event. Therefore, the video likely presents false information. \\
Prediction & Fake \\
\midrule
\textbf{Final Prediction} & Fake\\
\midrule
\textbf{Label} & Fake \\
\bottomrule
\end{tabular}
}
\label{Tab:CaseFakeTT}
\end{table}

\begin{figure}[!t]
\centering 
\includegraphics[scale=1.8]{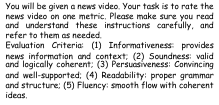}
\caption{Detailed prompts for G-Eval framework.}
\label{fig:G_eval_prompt}
\end{figure}

\section{More Model Explainability}
\label{APP:MOREEXP}

\subsection{Detail Settings of G-Eval}
\label{G_eval}
To ensure reproducibility, we clarify that we employed GPT-4o as the evaluator model, and set the temperature to $0$ to minimize randomness and ensure consistent scoring. The evaluation framework and prompt were adapted from the standard G-Eval framework \cite{liu2023g}, instructing the model to rate explanations on a 1-5 Likert scale based on the specific metric definitions. The prompt of the G-Eval framework can be found in Fig.~\ref{fig:G_eval_prompt}.

\subsection{Qualitative Analysis on Explainability} 
\label{APP:Qualitative}

To demonstrate the explainability of our proposed NVKE-CEI framework, we present a representative fake news video case from the FakeTT dataset in Table~\ref{Tab:CaseFakeTT}. This case illustrates how NVKE-CEI's dual-perspective fact-checking mechanism generates comprehensive, evidence-grounded explanations for veracity predictions. In this example, the video falsely claims the moon was struck by an asteroid. The content-based LLM identifies key red flags: OCR text contradicts astronomical plausibility, visual elements show computer-generated effects inconsistent with authentic imagery, and sensational hashtags (\#scary) suggest intent to misinform rather than inform. Simultaneously, the evidence-based LLM generates targeted queries (e.g., ``\textit{Moon struck by asteroid October 2023}'') and finds no credible sources corroborating the claim—only discussions of potential future events. 
The judge model integrates the two streams to predict \texttt{fake}, while the content-based reasoning and the complete sequence of evidence-based reasoning processes constitute the content-grounded explanation $e$.
We present case studies for error analysis in the Appendix~\ref{Error}.

\begin{table*}[!t]
\centering
\caption{Error Cases.}
\resizebox{0.83\textwidth}{!}{
\begin{tabular}{>{\centering\arraybackslash}cm{15mm}m{72mm}m{72mm}}
\toprule
 & & \textbf{FakeSV} & \textbf{FakeTT} \\
 \midrule
 \textbf{News Video} & & \begin{minipage}[b]{0.33\columnwidth}
            \centering
            \includegraphics[width=\linewidth]{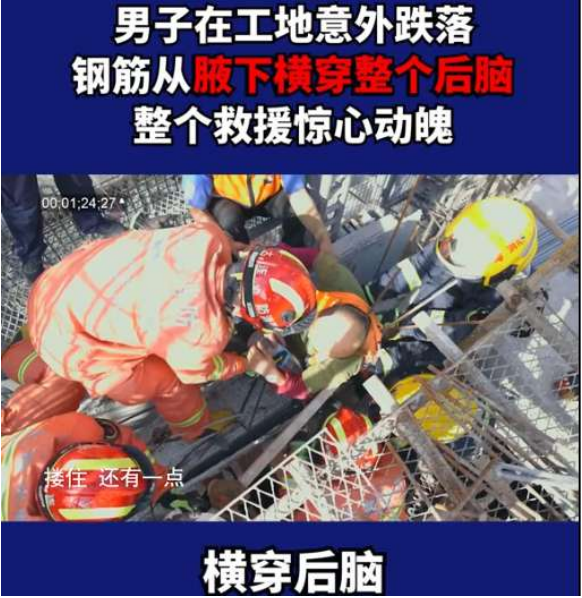}
            \end{minipage} & \begin{minipage}[b]{0.35\columnwidth}
            \centering
            \includegraphics[width=\linewidth]{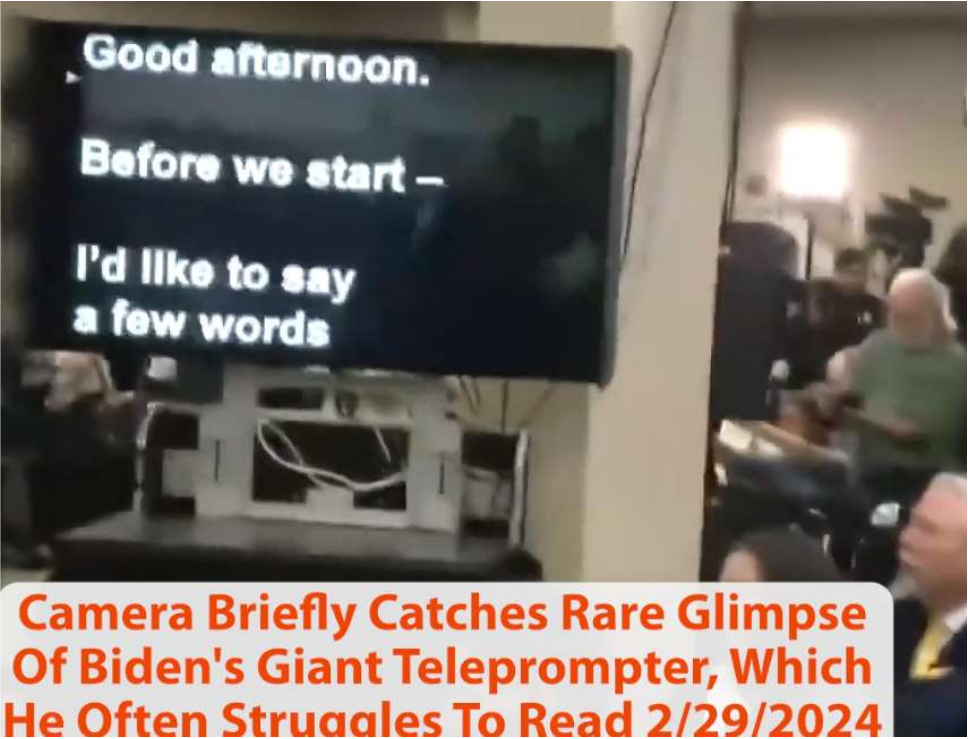}
            \end{minipage}\\
\textbf{Corresponding Text} & & Man falls at a construction site, a steel rod impales him from armpit to back of head; the rescue was dramatic. & ``\#HYGONews'' ``\#gop'' ``\#fyp'' ``\#JoeBiden'' \\
\textbf{News Summary} & & The video reports an accident where a man falls at a construction site, with a steel rod piercing from his armpit through the back of his head. It describes a thrilling rescue operation. & The news video claims to show instances where President Joe Biden struggled with reading a giant teleprompter, spoke casually to a government employee, and seemed confused about his location during a border briefing. These events are suggested to have happened on February 29, 2024. \\
\midrule
\textbf{Content-based LLM} & Reasoning Process & The video claims a man was impaled by rebar at a construction site, a highly improbable scenario. It relies only on text, not actual footage, likely exaggerating to attract viewers. The mention of a ``second episode'' suggests fabricated content for viewer engagement. This could be considered fake news, using sensationalism and inaccuracies to gain attention. & The video shows text information that implies unusual behavior from President Joe Biden, such as struggling to read a teleprompter or asking ``Where am I?'', and citing a date of February 29, 2024, which is nonexistent since February 2024 only has 28 days. This implies the narrative might not be true. Additionally, the lack of synchronization between the video and typical behaviors expected from a public figure raises doubts about the authenticity. The combination indicates that the video has likely been manipulated or misrepresented. Thus, it is fake news. \\
& Prediction & Fake & Fake \\ 
\midrule
\multirow{8}{*}{\textbf{Evidence-based LLM}} & Query Generation & & \\
& Reasoning Process & The video describes a severe incident involving personal safety, which would typically be reported in the news. Many online videos use exaggerated content to attract views, so it's essential to verify such incidents with credible news sources. Next, I should use search tools to find any real news reports that match this description. & The video claims to document President Biden's struggles with a teleprompter, confusion at a border briefing, and a hot mic moment with a government employee. To judge the authenticity, I need to verify these events with credible sources. I will perform a web search using the suggested queries. \\
& Queries & ``\textit{Real Case of Rebar Impalement at Construction Site}'', ``\textit{Rebar Through the Brain Rescue}'', ``\textit{Construction Site Rebar Accident News}'' & ``\textit{Biden teleprompter struggles February 2024}'', ``\textit{Biden border briefing confusion February 2024}'', ``\textit{Biden hot mic comments February 2024}''  \\
\cmidrule{2-4}
& Query Re-generation & & \\
& Reasoning Process & No need to regenerate queries. & No need to regenerate queries. \\
& Re-generate Queries & None & None \\
\cmidrule{2-4}
 & RAG & \multicolumn{2}{c}{\textit{Omitted for space}} \\
\cmidrule{2-4}
& Veracity Prediction &  & \\
& Reasoning Process & Search results found no credible news matching the video's description, suggesting the incident's authenticity is questionable. The lack of mainstream media coverage and the video's source indicate potential exaggeration or false content. & The search results do not provide evidence supporting the claims made in the video for the date mentioned (February 29, 2024). Although Biden has had gaffes with teleprompters and hot mic incidents in the past, the specific incidents on the given date are not corroborated by any credible sources. This discrepancy suggests the video may be misleading or fabricated. \\
& Prediction & Fake & Fake \\
\midrule
& Final Prediction & Fake & Fake \\ 
\midrule
\textbf{Label} & & Real & Real \\
\bottomrule
\end{tabular}
}
\label{T4}
\end{table*}

\section{Error Analysis}
\label{Error}

We present one error case from each of the FakeSV and FakeTT datasets in Table~\ref{T4}. The failed predictions reveal limitations in NVKE-CEI's dual-perspective fact-checking when both LLM checkers converge on incorrect conclusions.

\vpara{Convergent Failure of Dual Fact-Checkers.} In both FakeSV and FakeTT cases, content-based and evidence-based LLMs independently predicted fake despite the ground truth being real. When both checkers fail simultaneously, the veracity score fusion module lacks corrective signals, leading to final prediction errors. This suggests the dual-perspective approach, while generally effective, may amplify rather than mitigate errors in certain edge cases.

\vpara{Content-based LLM's Over-reliance on Perceived Plausibility.} The content-based LLM exhibits heightened skepticism toward unusual but genuine events. In FakeSV, it flagged the construction accident as fake because it deemed the rebar impalement \textit{highly improbable}. This suggests the four-dimensional analysis may disadvantage rare but authentic events lacking conventional visual documentation. 
For FakeTT, the content-based LLM incorrectly treated February 29, 2024 as nonexistent, and this factual reasoning error contributed to the false-positive prediction.

\vpara{Evidence-based LLM's Explanations of Negative Evidence.} The evidence-based LLM treated absence of evidence as evidence of absence. For FakeSV, failed query results led to concluding fabrication, overlooking that: (1) incidents may occur in regions with limited English reporting; (2) local outlets may use different terminology; (3) temporal restrictions may exclude relevant sources. The ReAct framework did not trigger additional iterations, prematurely determining it had sufficient evidence.

\begin{figure*}[!t]
\centering 
\includegraphics[scale=1.7]{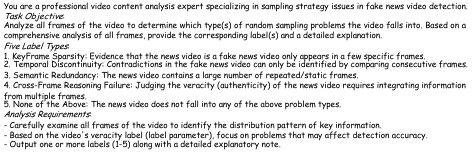}
\caption{Detailed prompts for pilot study.}
\label{fig:pilotstudy_prompt}
\end{figure*}

\begin{figure*}[!t]
\centering 
\includegraphics[scale=1.7]{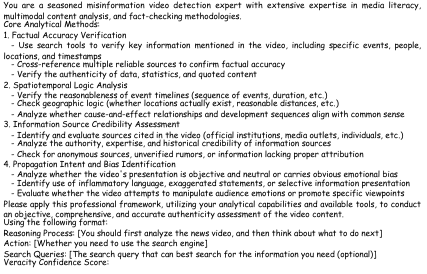}
\caption{Detailed prompts for evidence-based LLM.}
\label{fig:evidencebasedllm_prompt}
\end{figure*}

\begin{figure*}[!t]
\centering 
\includegraphics[scale=1.7]{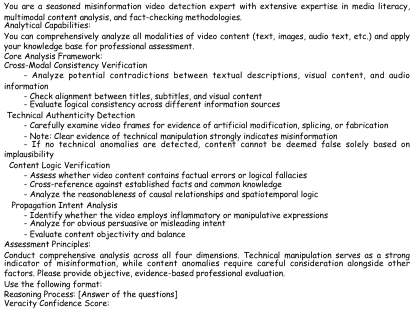}
\caption{Detailed prompts for content-based LLM.}
\label{fig:contentbasedllm_prompt}
\end{figure*}

\begin{figure*}[!t]
\centering 
\includegraphics[scale=1.7]{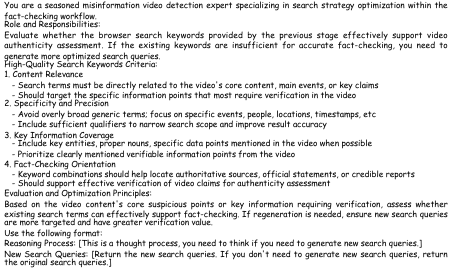}
\caption{Detailed prompts for query regeneration.}
\label{fig:query_regeneration_prompt}
\end{figure*}

\section{Prompts for Pilot Study and NVKE-CEI}

\subsection{Prompt for Pilot Study}
\label{P:Pilot}

The detailed prompt for the pilot study is presented in Fig.~\ref{fig:pilotstudy_prompt}.

\subsection{Prompt for Evidence-based LLM}
\label{P:Evidence}

The detailed prompt for the evidence-based LLM is presented in Fig.~\ref{fig:evidencebasedllm_prompt}.

\subsection{Prompt for Content-based LLM}
\label{P:Content}

The detailed prompt for the content-based LLM is presented in Fig.~\ref{fig:contentbasedllm_prompt}.

\subsection{Prompt for Query Regeneration}
\label{P:QueryRegen}

The detailed prompt for the query regeneration is presented in Fig.~\ref{fig:query_regeneration_prompt}.

\section{More Related Work}
\label{MoreRW}

\vpara{Text-image-based Fake News Detection.} 
Visual modalities have proven effective in fake news detection \cite{cao2020exploring, tufchi2023comprehensive, alam2021survey}.
Previous studies on text-image-based fake news detection can be categorized into supervised training-based and evidence LLM-based methods.
Supervised training-based methods \cite{jin2017multimodal, khattar2019mvae, chen2022cross} focus on learning correlations between text and image modalities, employing techniques such as attention-based RNNs \cite{jin2017multimodal} or Variational Autoencoders \cite{khattar2019mvae}.
Evidence LLM-based methods leverage LLMs' reasoning abilities for fact-checking \cite{hu2023large, chen2023can, wang2024mfc}. However, LLM hallucination remains problematic \cite{ji2023towards, perkovic2024hallucinations}. Recent methods \cite{wang2023explainable, liu2025detect, singhal2024evidence, khaliq2024ragar} incorporate RAG by extracting queries from news content and retrieving web information through search engines. Yet these methods face two critical limitations: generated queries may contain redundant information or omit specific keywords \cite{cuconasu2024power, chen2024benchmarking, xiang2024certifiably}, and the retrieved evidence may contain noisy information that compromises detection performance.

\vpara{Explainable Fake News Detection.}
Research on explainable fake news video detection remains limited, with only a few studies \cite{wu2024interpretable, chen2025multimodal} addressing this issue. \citet{wu2024interpretable} propose using attention backtracking to highlight suspicious text within the corresponding text of the news video. However, this approach merely identifies high-contribution regions for veracity prediction rather than providing the news content-grounded explanations. \citet{chen2025multimodal} propose MRGT to generate explanations exclusively for fake news videos while omitting the explanation process for real news videos. Moreover, MRGT requires both labeled news data and human-generated explanations for training.
Beyond fake news video detection, several studies have attempted to generate human-readable explanations for text-image-based fake news detection \cite{wang2023explainable, khaliq2024ragar, choi2024fact, singhal2024evidence}. 
\citet{wang2024explainable} utilize comment information for fake news debunking, while \citet{khaliq2024ragar} employ RAG to gather evidence from the web and generate explanations using LLM.

\end{document}